\documentclass{llncs}
\usepackage{amssymb,amsmath,subfigure,graphicx, pxfonts}
\usepackage{array}
\usepackage{rotating}
\usepackage{lmodern,blindtext}
\usepackage{epsfig}
\usepackage{multirow}
\usepackage{qtree}
\usepackage{stmaryrd}
\usepackage{ctable}

\begin{document}
\title{Comparative Study of View Update Algorithms in Rational Choice Theory \thanks{This paper extends work from Delhibabu [34] and Mayol [83]}}
\author{Radhakrishnan Delhibabu}
\institute{Informatik 5, Knowledge-Based Systems Group\\
RWTH Aachen, Germany\\
\email{delhibabu@kbsg.rwth-aachen.de}
} \maketitle
\begin{abstract}
The dynamics of belief and knowledge is one of the major components
of any autonomous system  that should be able to incorporate new
pieces of information. We show that knowledge base dynamics has
interesting connection with kernel change via hitting set and
abduction. The approach extends and integrates standard techniques
for efficient query answering and integrity checking. The generation
of hitting set is carried out through a hyper tableaux calculus and
magic set that is focused on the goal of minimality. Many different
view update algorithms have been proposed in the literature to
address this problem. The present paper provides a comparative study
of view update algorithms in rational approach.

\vspace{0.5cm}

\end{abstract}
\section{Introduction}
Modeling intelligent agents' reasoning requires designing knowledge
bases for the purpose of performing symbolic reasoning. Among the
different types of knowledge representations in the domain of
artificial intelligence, logical representations stem from classical
logic. However, this is not suitable for representing or treating
items of information containing vagueness, incompleteness or
uncertainty, or knowledge base evolution that leads the agent to
change his beliefs about the world.

When a new item of information is added to a knowledge base, it may
become inconsistent. In the argumentation theory people trying to
solve the same principle \cite{Amg,Bent1,Bent2,Bent3} in different
framework. Revision means modifying the knowledge base in order to
maintain consistency \cite{Mart}, while keeping the new information
and removing (contraction) or not removing the least possible
previous information. In our case, update means revision and
contraction, that is insertion and deletion in database perspective.
Previous work \cite{Arav1,Arav} makes connections with contraction
from knowledge base dynamics.

Our knowledge base dynamics is defined in two parts: an immutable
part (formulae) and updatable part (literals) (for definition and
properties see works of Nebel \cite{Nebel} and Segerberg
\cite{Seg}). Knowledge bases have a set of integrity constraints
(see the definitions in later section). In the case of finite
knowledge bases, it is sometimes hard to see how the update
relations should be modified to accomplish certain knowledge base
updates.

\begin{example} Consider a database with an (immutable) rule that a
staff member is a person who is currently working in a research
group under a chair. Additional (updatable) facts are that matthias
and gerhard are group chairs, and delhibabu and aravindan are staff
members in group info1. Our first integrity constraint (IC) is that
each research group has only one chair ie. $\forall x,y,z$ (y=x)
$\leftarrow$ group\_chair(x,y) $\wedge$ group\_chair(x,z). Second
integrity constraint is that a person can be a chair for only one
research group ie. $\forall x,y,z$ (y=z)$\leftarrow$
group\_chair(y,x) $\wedge$ group\_chair(z,x).

\begin{center}
\underline {Immutable part}: staff\_chair(X,Y)$\leftarrow$
staff\_group(X,Z),group\_chair(Z,Y). \vspace{0.5cm}

\underline{Updatable part}: group\_chair(infor1,matthias)$\leftarrow$ \\
\hspace{2.4cm}group\_chair(infor2,gerhard)$\leftarrow$ \\
\hspace{2.6cm}staff\_group(delhibabu,infor1)$\leftarrow$ \\
\hspace{2.6cm}staff\_group(aravindan,infor1)$\leftarrow$ \\
\end{center}
Suppose we want to update this database with the information,
staff\_chair(aravin\-dan,gerhard); From the immutable part, we can
deduce that this can be achieved by asserting
staff\_group(\underline{aravindan},Z) $\bigwedge$
group\_chair(Z,\-\underline{gerhard})

\end{example}

If we are restricted to definite clauses, there are three plausible
ways to do this: first case is, aravindan and gerhard belong to
infor1, i.e, staff\_group(\underline{aravindan},\-info1) $\bigwedge$
group\_chair\-(info1,\underline{gerhard}). We need to delete both
base facts group\_chair\-(infor1,matthias)$\leftarrow$ and
group\_chair(infor2,gerhard)$\leftarrow$, because our first IC as
well as second IC would be violated otherwise. In order to change
the view, we need to insert group\_chair(infor1,gerhard)$\leftarrow$
as a base fact. Assume that we have an algorithm that deletes the
base facts staff\_group(delhibabu,infor1)$\leftarrow$ from the
database. But, no rational person will agree with such an algorithm,
because the fact staff\_group(delhibabu,infor1)$\leftarrow$ is not
"relevant" to the view atom.

Second case, aravindan and gerhard belong to infor2, that is
staff\_group(\underline{aravin}\-\underline{dan},info2) $\bigwedge$
group\_chair(info2,\underline{gerhard}). Simply, insert the new fact
staff\_group(ara\-vindan,infor2)$\leftarrow$ to change the view.
Suppose an algorithm deletes the base facts
staff\_group(aravind\-an,infor1)$\leftarrow$ from the database, then
it can not be "rational" since these facts are not "relevant" to the
view atom.

Third case, aravindan and gerhard belong to infor3 (free assignment
of the group value), that is
staff\_group(\underline{aravindan},info3) $\bigwedge$
group\_chair(info3,\underline{gerhard}). Suppose, we insert new base
fact group\_chair(info3,gerhard) $\leftarrow$, our second IC does
not follow. Suppose an algorithm inserts the new base fact
staff\_group(aravin\-dan,infor2)$\leftarrow$ or
staff\_group(aravindan,infor1)$\leftarrow$ is deleted, then it can
not be "rational".

The above example highlights the need for some kind of "relevance
policy" to be adopted when a view atom is to be inserted to a
deductive database. How many such axioms and policies do we need to
characterize a "good" view update? When are we sure that our
algorithm for view update is "rational"? Clearly, there is a need
for an axiomatic characterization of view updates. By axiomatic
characterization, we mean explicitly listing all the rationality
axioms that are to be satisfied by any algorithm for view update.

When dealing with the revision of a knowledge base (both insertions
and deletions), there are other ways to change a knowledge base and
it has to be performed automatically also. Considering the
information, change is precious and must be preserved as much as
possible. The \emph{principle of minimal change} \cite{Herz,Schul}
can provide a reasonable strategy. On the other hand, practical
implementations have to handle contradictory, uncertain, or
imprecise information, so several problems can arise: how to define
efficient change in the style of Carlos Alchourr$\acute{o}$n, Peter
G$\ddot{a}$rdenfors, and David Makinson (AGM) \cite{Alch}; what
result has to be chosen \cite{Lak,Lobo,Nayak1}; and finally,
according to a practical point of view, what computational model to
support for knowledge base revision has to be provided?

The basic idea in \cite{Beh,Arav2} is to employ the model generation
property of hyper tableaux and magic set to generate models, and
read off diagnosis from them. One specific feature of this diagnosis
algorithm is the use of semantics (by transforming the system
description and the observation using an initial model of the
correctly working system) in guiding the search for a diagnosis.
This semantical guidance by program transformation turns out to be
useful for database updates as well. More specifically we use a
(least) Herbrand model of the given database to transform it along
with the update request into a disjunctive logic program in such a
way that the models of this transformed program stand for possible
updates. This paper aims at studying the view update algorithms in
relational databases. First, we define a framework for highlighting
the basic theory of minimal change. Thus, we present a generalized
revision algorithm based on abductive explanation for knowledge base
revision and main view update method.

The rest of paper is organized as follows: First we start with
preliminaries in Section 2. In Section 3, we introduce knowledge
base dynamics along with the concept of generalized revision, and
revision operator for knowledge base. Section 4 studies the
relationship between knowledge base dynamics and abduction. We
discuss an important application of knowledge base dynamics in
providing an axiomatic characterization for updating view literal to
databases. We briefly discuss hyper tableaux calculus and magic set
in Section 5. We present two variants of our rational and efficient
algorithm for view update in Section 6. In Section 7, we discuses
six basic dimensions in the process of view updating and comparative
study of view update algorithms in rational approach is presented.
In Section 7, we give brief overview. In Section 8 we draw
conclusions with a summary of our contribution and indicate future
directions of our investigation. All proofs can be found in the
Appendix.

\section{Preliminaries}

We consider a propositional language $\mathcal{L_P}$ defined from a
finite set of propositional variables $\mathcal{P}$ and the standard
connectives. We use lower case Roman letters $a, b, x, y,...$ to
range over elementary letters and Greek letters $\varphi, \phi,
\psi, ...$ for propositional formulae. Sets of formulae are denoted
by upper case Roman letters $A,B, F,K, ....$. A literal is an atom
(positive literal), or a negation of an atom (negative literal).

For any formula $\varphi$, we write $E(\varphi)$ to mean the set of
the elementary letters that occur in $\varphi$. The same notation
also applies to a set of formulae. For any set $F$ of formulae,
$L(F)$ represents the sub-language generated by $E(F)$, i.e., the
set of all formulae $\varphi$ with $E(\varphi) \subseteq E(F)$.

Horn formulae are defined \cite{Delg} as follows:
\begin{enumerate}
\item[1.] Every $a \in \Phi$ where $\Phi \in \mathcal{L_P} \cup \{ \bot \}$ , $a$ and $\neg a$ are Horn clauses.
\item[2.] $a \leftarrow a_1 \land a_2 \land ... \land a_n$ is a Horn clause, where $n \geq 0$ and
$a, a_i \in \Phi$ ($1 \leq i \leq n$).
\item[3.] Every Horn clause is a Horn formula, $a$ is called head and $a_i$
is body of the Horn formula.
\item[4.] If $\varphi$ and $\psi$ are Horn formulae, so is $\varphi\land \psi$.
\end{enumerate}

A definite Horn clause is a finite set of literals (atoms) that
contains exactly one positive literal which is called the head of
the clause. The set of negative literals of this definite Horn
clause is called the body of the clause. A Horn clause is
non-recursive, if the head literal does not occur in its body. We
usually denote a Horn clause as head$\leftarrow$body. Let
$\mathcal{L_H}$ be the set of all Horn formulae with respect to
$\mathcal{L_P}$. A formula $\phi$ is a syntactic consequence within
$\mathcal{L_P}$ of a set $\Gamma$ of formulas if there is a formal
proof in $\mathcal{L_P}$ of $\phi$ from the set $\Gamma$ is $\Gamma
\vdash_{\mathcal{L_P}} \phi$.

A immutable part is a function-free clause of the form $a \leftarrow
a_1 \land a_2 \land ... \land a_n$, with $n\geq 1$ where $a$ is an
atom denoting the immutable part's head and $a_1 \land a_2 \land ...
\land a_n$ are literals. i.e., positive or negative atoms,
representing the body of the Horn clauses.

Formally, a finite Horn knowledge base $KB$ is defined as a finite
set of formulae from language $\mathcal{L_{H}}$, and divided into
three parts: an immutable theory $KB_{I}$ is a Horn formula
(head$\leftarrow$body), which is the fixed part of the knowledge;
updatable theory $KB_{U}$ is a Horn clause (head$\leftarrow$); and
integrity constraint $KB_{IC}$ representing a set of clauses
($\leftarrow$body).

\begin{definition} [Horn Knowledge Base] \label{D1} A Horn knowledge base, KB
is a finite set of Horn formulae from language $\mathcal{L_{H}}$,
s.t $KB=KB_{I}\cup KB_{U}\cup KB_{IC}$, $KB_{I}\cap
KB_{U}=\varnothing$ and $KB_{U}\cap KB_{IC}=\varnothing$.
\end{definition}

In the AGM framework, a belief set is represented by a deductively
closed set of propositional formulae. While such sets are infinite,
they can always be finitely representable. However, working with
deductively closed, infinite belief sets is not very attractive from
a computational point of view. The AGM approach to belief dynamics
is very attractive in its capturing the rationality of change, but
it is not always easy to implement either Horn formula based partial
meet revision. In real application from artificial intelligence and
databases, what is required is to represent the knowledge using a
finite Horn knowledge base. Further, a certain part of the knowledge
is treated as immutable and should not be changed.

Knowledge base change deals with situations in which an agent has to
modify its beliefs about the world, usually due to new or previously
unknown incoming information, also represented as formulae of the
language. Common operations of interest in Horn knowledge base
change are the expansion of an agent's current Horn knowledge base
KB by a given Horn clause $\varphi$ (usually denoted as
KB+$\varphi$), where the basic idea is to add regardless of the
consequences, and the revision of its current beliefs by $\varphi$
(denoted as KB * $\varphi$), where the intuition is to incorporate
$\varphi$ into the current beliefs in some way while ensuring
consistency of the resulting theory at the same time. Perhaps the
most basic operation in Horn knowledge base change, like belief
change, is that of contraction (AGM \cite{Alch}), which is intended
to represent situations in which an agent has to give up $\varphi$
from its current stock of beliefs (denoted as KB-$\varphi$).

\begin{definition} [AGM Contraction] Let KB be a Horn knowledge base, and $\alpha$ a belief
that is present in KB. Then \emph{contraction} of KB by $\alpha$,
denoted as $KB-\alpha$, is a consistent belief set that excludes
$\alpha$
\end{definition}

\begin{definition} [Levi Identity] \label{D2} Let - be an AGM contraction
operator for KB. A way to define a revision is by using Generalized
Levi Identity:
\begin{center}
$KB*\alpha~=~(KB-\neg\alpha)\cup\alpha$
\end{center}
\end{definition}

Then, the revision can be trivially achieved by expansion, and the
axiomatic characterization could be straightforwardly obtained from
the corresponding characterizations of the traditional models
\cite{Fal}. The aim of our work is not to define revision from
contraction, but rather to construct and axiomatically characterize
revision operators in a direct way.


\section{Knowledge base dynamics}

AGM \cite{Alch} proposed a formal framework in which
revision(contraction) is interpreted as belief change. Focusing on
the logical structure of beliefs, they formulate eight postulates
which a revision knowledge base (contraction knowledge base was
discussed in \cite{Arav}) has to verify.

In the AGM approach, a belief is represented by a sentence over a
suitable language $\mathcal{L_{H}}$, and a belief $KB$ is
represented by a set of sentence that are close wrt the logical
closure operator Cn. It is assumed that $\mathcal{L_{H}}$, is closed
under application of the boolean operators negation, conjunction,
disjunction, and implication.

\begin{definition} \label{D9} Let KB be a knowledge base with an immutable part
$KB_{I}$. Let $\alpha$ and $\beta$ be any two clauses from
$\mathcal{L_H}$. Then, $\alpha$ and $\beta$ are said to be
\emph{KB-equivalent} iff the following condition is satisfied:
$\forall$ set of Horn clauses E $\subseteq \mathcal{L_H}$:
$KB_{I}\cup E\vdash\alpha$ iff $KB_{I}\cup E\vdash\beta$.
\end{definition}

These postulates stem from three main principles: the new item of
information has to appear in the revised knowledge base, the revised
base has to be consistent and revision operation has to change the
least possible beliefs. Now we consider the revision of a Horn
clause $\alpha$ wrt KB, written as $KB*\alpha$. The rationality
postulates for revising $\alpha$ from KB can be formulated as
follows:

\begin{definition} [Rationality postulates for knowledge base
revision] \label{10}
\begin{enumerate}
\item[]\hspace{-0.6cm}(KB*1)\hspace{0.2cm}  \emph{Closure:} $KB*\alpha$ is a knowledge base.
\item[]\hspace{-0.6cm}(KB*2)\hspace{0.2cm}  \emph{Weak Success:} if $\alpha$ is consistent with $KB_{I}\cup KB_{IC}$ then
$\alpha \subseteq KB*\alpha$.
\item[]\hspace{-0.6cm}(KB*3.1)  \emph{Inclusion:} $KB*\alpha\subseteq
Cn(KB\cup\alpha)$.
\item[]\hspace{-0.6cm}(KB*3.2)  \emph{Immutable-inclusion:} $KB_{I}\subseteq
Cn(KB*\alpha)$.
\item[]\hspace{-0.6cm}(KB*4.1)  \emph{Vacuity 1:} if $\alpha$ is
inconsistent with $KB_{I}\cup KB_{IC}$ then $KB*\alpha=KB$.
\item[]\hspace{-0.6cm}(KB*4.2)  \emph{Vacuity 2:} if $KB\cup \alpha \nvdash \perp$ then $KB*\alpha$ = $KB \cup
\alpha$.
\item[]\hspace{-0.6cm}(KB*5)\hspace{0.3cm}   \emph{Consistency:} if $\alpha$ is consistent with $KB_{I}\cup KB_{IC}$
 then $KB*\alpha$ is consistent with $KB_{I}\cup KB_{IC}$.
\item[]\hspace{-0.6cm}(KB*6)  \hspace{0.2cm} \emph{Preservation:} If $\alpha$ and $\beta$ are
KB-equivalent, then $KB*\alpha \leftrightarrow KB*\beta$.
\item[]\hspace{-0.6cm}(KB*7.1)  \emph{Strong relevance:} $KB*\alpha\vdash \alpha$ If $KB_{I}\nvdash\neg\alpha$
\item[]\hspace{-0.6cm}(KB*7.2)  \emph{Relevance:} If $\beta\in KB\backslash KB*\alpha$,
then there is a set $KB'$ such that\\ $KB*\alpha\subseteq
KB'\subseteq KB\cup\alpha$, $KB'$ is consistent $KB_{I}\cup KB_{IC}$
with $\alpha$, but $KB' \cup \{\beta\}$ is inconsistent $KB_{I}\cup
KB_{IC}$ with $\alpha$.
\item[]\hspace{-0.6cm}(KB*7.3)  \emph{Weak relevance:} If $\beta\in KB\backslash KB*\alpha$,
then there is a set $KB'$ such that $KB'\subseteq KB\cup\alpha$,
$KB'$ is consistent $KB_{I}\cup KB_{IC}$ with $\alpha$, but $KB'
\cup \{\beta\}$ is inconsistent $KB_{I}\cup KB_{IC}$ with $\alpha$.
\end{enumerate}
\end{definition}

To revise $\alpha$ from KB, only those information that are relevant
to $\alpha$ in some sense can be added (as the example from the
introduction illustrates). $(KB*7.1)$  is a very strong axiom
allowing only minimum changes, and certain rational revisions can
not be carried out. So, relaxing this condition (example with more
details can be found in \cite{Arav}) allows for weakening strong
relevance to relevance only. The relevance policy $(KB*7.2)$,
however, still does not permit rational revisions, so we need to go
one step further. With $(KB*7.3)$ the relevance axiom is further
weakened and the resulting conditions are referred to as
"core-retainment".


\subsection{Relationship with Abductive Logic Grammars}
The relationship between Horn knowledge base dynamics and abduction
was introduced by the philosopher Pierce (see \cite{Alis}). We show
how abduction grammar could be used to realize revision with
immutability condition. A special subset of literal (atoms) of
language $\mathcal{L_{H}}$, \emph{abducibles} Ab, are designated for
abductive reasoning. Our work is based on atoms, so we combine
Christiansen and Dahl \cite{Chris} grammars method to our theory.

\begin{definition}[Abductive grammar] \label{D18} An abductive grammar $\Gamma$ is a 6-tuple $\langle
N,T,IC,KB,\-R,S\rangle$ where

\begin{enumerate}
\item[-] N are nonterminal symbols in immutable part ($KB_I$).
\item[-] T is a set of terminal symbols in updatable part ($KB_U$).
\item[-] IC is the integrity constraint to Horn knowledge base ($KB_{IC}$).
\item[-] KB is the Horn knowledge base which consists of $KB=KB_{I}\cup KB_{U}\cup KB_{IC}$.
\item[-] R is a set of rules, $R\subseteq KB$.
\item[-] S is the revision of literals (atoms), called the start symbol.
\end{enumerate}
\end{definition}

\begin{definition}[Constraint system] \label{D19}
A constraint system for a abduction is a pair $\langle
KB^{Ab},\-KB^{BG} \rangle$, where $KB^{Ab}(\Delta)$ is a set of
propositions (abducibles) and $KB^{BG}$ background Horn knowledge
base.
\end{definition}

\textbf{Notations:} From grammar point, $KB^{BG}$ is set all Horn
formulae from R and $KB^{Ab}$ is set of abducibles from T.

\begin{definition}[Minimal abductive explanation] \label{D20} Let KB be a Horn knowledge base and $\alpha$ an
observation to be explained. Then, for a set of abducibles
$(KB^{Ab})$, $\Delta$ is said to be an abductive explanation wrt
$KB^{BG}$ iff $KB^{BG}\cup \Delta\vdash \alpha$. $\Delta$ is said to
be \emph{minimal} wrt $KB^{BG}$ iff no proper subset of $\Delta$ is
an abductive explanation for $\alpha$, i.e. $\nexists\Delta^{'}$
s.t. $KB^{BG}\cup\Delta^{'}\vdash\alpha$.
\end{definition}

Since an incision function is adding and removing only updatable
elements from each member of the kernel set, to compute a
generalized revision of $\alpha$ from KB, we need to compute only
the abduction in every $\alpha$-kernel of KB. So, it is now
necessary to characterize precisely the abducibles present in every
$\alpha$-kernel of KB. The notion of minimal abductive explanation
is not enough to capture this, and we introduce locally minimal and
KB-closed abductive explanations.

\begin{definition}[Local minimal abductive explanations] \label{D21}
Let $KB^{BG'}$ be a smallest subset of $KB^{BG}$, s.t $\Delta$ is a
minimal abductive explanation of $\alpha$ wrt $KB^{BG'}$ (for some
$\Delta$). Then $\Delta$ is called local minimal for $\alpha$ wrt
$KB^{BG}$.
\end{definition}

\begin{definition} [Constraint abduction system] A constrained abductive grammar is a pair
$\langle \Gamma,C\rangle$, where $\Gamma$ is an abductive grammar
and $C$ a constraint system for abduction, $\Gamma$=$\langle N,T,R,S
\rangle$ and C=$\langle KB^{BG},KB^{Ab},IC\rangle$.
\end{definition}

Given a constrained abductive grammar $\langle \Gamma,C\rangle$ as
above, the constrained abductive recognition problem for $\tau \in
T^*$ is the problem of finding an admissible and denial knowledge
base from $KB^{Ab}$ and  such that $\tau \in
\mathcal{L}_{P}(\Gamma_{KB^{Ab}})$ where
$\mathcal{L}_{P}(\Gamma_{KB^{Ab}})$ is propositional language over
abducibles in $\Gamma$, where $\Gamma_{KB^{Ab}}$ = $\langle N,T,
KB^{BG}\cup KB^{Ab},R,S \rangle$. In this case, $KB^{Ab}$ is called
a \emph{constrained (abductive) system of} $\tau$. Such that
$KB^{Ab}$ is minimal whenever no proper subset of it is an in $\tau$
given $\langle \Gamma,C\rangle$.

Let $KB^{Ab}\in(\{\Delta^{+},\Delta^{-}\})$. Here $\Delta^{+}$
refers to admission Horn knowledge base (positive atoms) and
$\Delta^{-}$ refers to denial Horn knowledge base(negative atoms)
wrt given $\alpha$. Then problem of abduction is to explain $\Delta$
with abducibles $(KB^{Ab})$, s.t.
$KB^{BG}\cup\Delta^{+}\cup\Delta^{-}\vdash \alpha$ and
$KB^{BG}\cup\Delta^{+}\models\alpha\cup\Delta^{-}$ are both
consistent with IC.

\begin{theorem} \label{T9}
Consider a constrained abductive grammar $AG = \langle
\Gamma,C\rangle$ with $\Gamma = \langle N,T,KB,R, S\rangle$ and $C =
\langle KB^{BG},KB^{Ab},IC\rangle$. Construct a abductive grammar
$\Delta(AG) =$$\langle N, T,KB^{BG},R,S\rangle$ by having, for any
$(\Delta^{+})$ $(or (\Delta^{-})$ from $KB^{Ab}$, the set of
acceptable results for accommodate$(\alpha,KB^{BG}\in\Delta^{+})$
being of the form $(KB^{Ab}\backslash \Delta^{+})$ where
($\Delta^{+}\in KB^{Ab'}$). $\Delta^{+}$ is a locally minimal set of
atoms (literals) $KB^{BG}\cup\Delta^{+}$ and $KB^{BG}\cup\Delta^{+}
\models \alpha$ is consistent with IC; if no such $(\Delta^{-})$
exists $($like denial $(\Delta^{-})$ being of the form
$(KB^{Ab}\backslash \Delta^{-})$. $\Delta^{-}$ is a locally minimal
set of atoms (literals) $KB^{BG}\cup\Delta^{-}$ and
$KB^{BG}\cup\Delta^{-} \models \alpha$ is consistent with IC$)$,
otherwise accommodate $(\alpha,KB^{BG}\in\Delta^{-})$ is not
possible.
\end{theorem}

Now, we need to connect the grammar system $\Gamma$ to the Horn
knowledge base $KB$, such that $KB_I\cup KB_U\cup
KB_{IC}=KB^{BG}\cup KB^{Ab}\cup IC$ holds. The connection between
locally minimal abductive explanation for $\alpha$ wrt $KB_I$ and
$\alpha$-kernel of KB, which is shown by the following lemma
immediately follows from their respective definitions.

\begin{lemma}
\hspace{0.5cm}
\begin{enumerate}
\item[1.] Let KB be a Horn knowledge base and $\alpha$ a Horn clause s.t.
$\nvdash\neg \alpha$. Let $\Delta^{+}$ and $\Delta^{-}$ be a
KB-closed locally minimal abductive explanation for $\alpha$ wrt
$KB_{I}$. Then, there exists a $\alpha$-kernel X of KB s.t. $X\cap
KB_{U}=\Delta^{+} \cup \Delta^{-}$.
\item[2.] Let KB be a Horn knowledge base and $\alpha$ a Horn clause s.t.
$\nvdash\neg \alpha$. Let X be a $\alpha$-kernel of KB and
$\Delta^{+} \cup \Delta^{-}=X\cap KB_{U}$. Then, $\Delta^{+}$ and
$\Delta^{-}$ are KB -locally minimal abductive explanations for
$\alpha$ wrt $KB_{I}$.
\end{enumerate}
\end{lemma}

An immediate consequence of the above lemma 4.1 is that it is enough
to compute all the KB-locally minimal abductive explanations for
$\alpha$ wrt $KB_{I}$ in order to revise $\alpha$ from KB. Thus, a
well-known abductive procedure to compute an abductive explanation
for $\alpha$ wrt $KB_I$ could be used.


\subsection{Generalized revision algorithm}

The problem of knowledge base revision is concerned with determining
how a request to change can be appropriately translated into one or
more atoms or literals.  In this section we develop a new
generalized revision algorithm. Note that it is enough to compute
all the KB-locally minimal abduction explanations for $\alpha$ wrt
$KB_I \cup KB_U \cup KB_{IC}$. If $\alpha$ is consistent with KB
then a well-known abductive procedure for compute an abductive
explanation for $\alpha$ wrt $KB_{I}$ could be used to compute kernel revision.\\

\begin{theorem} \label{T8} Let KB be a Horn knowledge base and $\alpha$
is formula.
\begin{enumerate}
  \item If Algorithm 1 produced KB'as a result of revising $\alpha$
  from KB, then KB' satisfies all the rationality postulates (KB*1) to
(KB*6) and (KB*7.3).
  \item Suppose $KB''$ satisfies all these rationality postulates
  for revising $\alpha$ from KB, then $KB''$ can be produced by Algorithm 1.
\end{enumerate}
\end{theorem}

$$\begin{array}{cc}\hline \text{\bf Algorithm 1} & \hspace{-4cm}
\text{\rm Generalized revision algorithm}\\\hline \text{\rm Input}:&
\hspace{-0.6cm}\text{\rm A Horn knowledge base}~ KB=KB_{I}\cup KB_{U}\cup KB_{IC}\\
&\text{\rm and a Horn clause}~
\alpha~ \text{\rm to be revised.}\\
\text{\rm Output:} & \text{\rm A new Horn knowledge base
}~KB'=KB_{I}\cup
KB_{U}^*\cup KB_{IC},\\
&\text{s.t.}~ KB'\text{\rm is
a generalized revision}~ \alpha~\text{\rm to KB.}\\
\text{\rm Procedure}~KB(KB,\alpha)&\\
\text{\rm begin}&\\
~~1.&\hspace{-0.5cm}\text{\rm Let V:=}~\{c\in KB_{IC}~|~ KB_I\cup
KB_{IC}~\text{\rm inconsistent
with}~\alpha~\text{\rm wrt}~c\}\\
&P:=N:=\emptyset~\text{\rm and}~KB'=KB\\
~~2.&\text{\rm While}~(V\neq \emptyset)\\
&\text{\rm select a subset}~V'\subseteq V\\
&\text{\rm For each}~v\in~V',~\text{\rm select a literal to be}\\
&\hspace{-0.1cm}\text{\rm remove (add to N) or a literal to be added (add to P) wrt KB}\\
&\text{\rm Let KB}~:=KR(KB,P,N)\\
&\hspace{-0.3cm}\text{\rm Let V:=}~\{c\in KB_{IC}~|~ KB_I~\text{\rm
inconsistent
with}~\alpha~\text{\rm wrt}~c\}\\
&\hspace{-0.7cm}\text{\rm return}\\
~~3.&\text{\rm Produce a new Horn knowledge base}~KB'\\
\text{\rm end.}&\\ \hline
\end{array}$$

\vspace{0.5cm}

$$\begin{array}{cc}\hline
\text{\bf Algorithm 2} \label{A2} &\\
\text{\rm Procedure}~
KR(KB,\Delta^{+},\Delta^{-})\\
\text{\rm begin}\\
1.~\text{\rm Let}~ P :=\{ e \in \Delta^{+} |~ KB_I\not\models e\}
~\text{\rm and}~ N :=\{ e \in \Delta^{-}
 |~KB_I\models e\}\\
2.~\text{\rm While}~(P\neq \emptyset)~\text{\rm or}~(N\neq \emptyset)\\
\text{\rm select a subset}~P'\subseteq P~ or ~N'\subseteq N \\
\hspace{-1.7cm}\text{\rm Construct a set}~S_1=\{X~|~X~\text{\rm is a
KB-closed
locally}\\
\text{\rm minimal abductive explanation wrt P} \}\\
\hspace{-1.7cm}\text{\rm Construct a set}~S_2=\{X~|~X~\text{\rm is a
KB-closed
locally}\\
\text{\rm  minimal abductive explanation wrt N } \}\\
\text{\rm Determine hitting set}~\sigma (S_1) \text{\rm ~and}~\sigma (S_2)\\
\hspace{-5.5cm}
\text{\rm If}~((N'=\emptyset)~and~(P'\neq\emptyset))\\
\hspace{-1cm}\text{\rm Produce}~KB'=KB_{I}\cup \{(KB_{U} \cup \sigma (S_1)\}\\
\hspace{-8.8cm}
\text{\rm else}\\
\text{\rm Produce}~KB'=KB_{I}\cup \{(KB_{U}\backslash
\sigma(S_2) \cup \sigma (S_1)\}\\
\hspace{-8.5cm}
\text{\rm end if}\\
\hspace{-5.5cm}
\text{\rm If}~((N'\neq\emptyset)~\text{\rm and}~(P'=\emptyset))\\
\hspace{-1.2cm}\text{\rm Produce}~KB'=KB_{I}\cup \{(KB_{U}\backslash
\sigma(S_2)\}\\
\hspace{-8.8cm}
\text{\rm else}\\
\text{\rm Produce}~KB'=KB_{I}\cup \{(KB_{U}\backslash
\sigma(S_2) \cup \sigma (S_1)\}\\
\hspace{-8.5cm}
\text{\rm end if}\\
 \text{\rm Let}~ P :=\{ e \in \Delta^{+} |~ KB_I\not\models e\}
~\text{\rm and}~ N :=\{ e \in \Delta^{-}
 |~KB_I\models e\}\\
3.~\text{\rm return}~ KB'\\
\text{\rm end.}\\ \hline
\end{array}$$


\section{Deductive database} A \it Deductive database \rm $DDB$ consists of three parts:
an \it intensional database \rm $IDB$ ($KB_I$), a set of definite
program clauses, \it extensional database \rm $EDB$ ($KB_U$), a set
of ground facts; and \it integrity constraints\rm ~$IC$. The
intuitive meaning of $DDB$ is provided by the \it Least Herbrand
model semantics \rm and all the inferences are carried out through
\it SLD-derivation. All the predicates that are defined in $IDB$ are
referred to as \emph{view predicates}  and those defined in $EDB$
are referred to as \emph{base predicates}. \rm Extending this
notion, an atom with a view predicate is said to be a \it view
atom,\rm and similarly an atom with base predicate is a \it base
atom. \rm Further we assume that $IDB$ does not contain any unit
clauses and no predicate defined in a given $DDB$ is both view and
base.

Two kinds of view updates can be carried out on a $DDB$: An atom,
that does not currently follow from $DDB$, can be \it inserted, \rm
or an atom, that currently follows from $DDB$ can be \it deleted.
 \rm When an atom $A$ is to be updated, the view update problem is to
insert or delete only some relevant $EDB$ facts, so that the
modified $EDB$ together with $IDB$ will satisfy the updating of $A$
to $DDB$.

Note that a $DDB$ can be considered as a knowledge base to be
revised. The $IDB$ is the immutable part of the knowledge base,
while the $EDB$ forms the updatable part. In general, it is assumed
that the language underlying a $DDB$ is fixed and the semantics of
$DDB$ is the least Herbrand model over this fixed language. We
assume that there are no function symbols implying that the Herbrand
Base is finite. Therefore, the $IDB$ is practically a shorthand of
its ground instantiation\footnotemark \footnotetext{a ground
instantiation of a definite program $P$ is the set of clauses
obtained by substituting terms in the Herbrand Universe for
variables in $P$ in all possible ways}  written as $IDB_G$. In the
sequel, technically we mean $IDB_G$ when we refer simply to $IDB$.
Thus, a $DDB$ represents a knowledge base where the immutable part
is given by $IDB_G$ and updatable part is the $EDB$. Hence, the
rationality postulates (KB*1)-(KB*6) and (KB*7.3) provide an
axiomatic characterization for update (insert and delete) a view
atom $A$ from a definite database $DDB$.

 Logic provides a conceptual level for understanding the meaning of relational databases.
 Hence, the rationality postulates (KB*1)-(KB*6) and
(KB*7.3) can provide an axiomatic characterization for view updates
in relational databases too. A relational database together with its
view definitions can be represented by a deductive database ($EDB$
representing tuples in the database and $IDB$ representing the view
definitions), and so the same algorithm can be used to delete view
extensions from relational deductive databases.

An update request U = B, where B is a set of base facts, is not true
in KB. Then, we need to find a transaction $T=T_{ins} \cup T_{del}$,
where $T_{ins} (\Delta_i)$ (resp. $T_{del}(\Delta_j)$) is the set of
facts, such that U is true in $DDB'=((EDB - T_{del} \cup T_{ins})
\cup IDB \cup IC)$. Since we consider stratifiable (definite)
deductive databases, SLD-tree can be used to compute the required
abductive explanations. The idea is to get all EDB facts used in a
SLD-derivation of $A$ wrt DDB, and construct that as an abductive
explanation for $A$ wrt $IDB_G$.

All solutions translate \cite{Mota} a view update request into a
\textbf{transaction combining insertions and deletions of base
relations} for satisfying the request. Further, a stratifiable
(definite) deductive database can be considered as a knowledge base,
and thus rationality postulates and insertion algorithm of the
previous section can be applied for view updates in database.


\section{View update method}

View updating \cite{Beh} aims at determining one or more base
relation updates such that all given update requests with respect to
derived relations are satisfied after the base updates have been
successfully applied.

\begin{definition}[View update] Let $DDB = \langle IDB,EDB,IC\rangle$ be a
stratifiable (definite) deductive database $DDB(D)$. A VU request
$\nu_{D}$ is a pair $\langle \nu^+_{D},\nu^-_{D}\rangle$ where
$\nu^+_{D}$ and $\nu^-_{D}$ are sets of ground atoms representing
the facts to be inserted into $D$ or deleted from $D$, resp., such
that $pred(\nu^+_{D}\cup \nu^-_{D}) \subseteq pred(IDB)$,
$\nu^+_{D}\cap \nu^-_{D} = \emptyset$, $\nu^+_{D}\cap
PM_{D}=\emptyset$ and $\nu^-_{D}\subseteq PM_{D}$.\rm
\end{definition}

Note that we consider again true view updates only, i.e., ground
atoms which are presently not derivable for atoms to be inserted, or
are derivable for atoms to be deleted, respectively. A method for
view updating determines sets of alternative updates satisfying a
given request. A set of updates leaving the given database
consistent after its execution is called \emph{VU realization}.

\begin{definition} [Induced update] Let $DDB = \langle IDB,EDB,
IC\rangle$ be a stratifiable (definite) deductive database and
$DDB=\nu_{D}$ a VU request. A VU realization is a base update
$u_{D}$ which leads to an induced update $u_{D\rightarrow D'}$ from
$D$ to $D'$ such that $\nu^+_{D}\subseteq PM_{D'}$ and
$\nu^-_{D}\cap PM_{D'}=\emptyset$.\rm
\end{definition}

There may be infinitely many realizations and even realizations of
infinite size which satisfy a given VU request. A breadth-first
search (BFS) is employed for determining a set of minimal
realizations $\tau_{D}= \{u^1_{D},\ldots, u^i_{D}\}$. Any $u^i_{D}$
is minimal in the sense that none of its updates can be removed
without losing the property of being a realization for $\nu_{D}$.

In \cite{Bau,Arav2} a variant of clausal normal form tableaux called
"hyper tableaux" is introduced. Since the hyper tableaux calculus
constitutes the basis for our view update algorithm, \it Clauses,
\rm i.e., multisets of literals, are usually written as the
disjunction $A_1\lor A_2\lor\cdots\lor A_m\lor~\text{not}~B_1\lor
~\text{not}~B_2\cdots\lor~\text{not}~B_n$ ($M\geq 0,n\geq 0$). The
literals $A_1,A_2,\ldots A_m$ (resp. $B_1,B_2,\ldots, B_n$) are
called the \it head (\rm resp. \it body) \rm of a clause. With
$\overline{L}$ we denote the complement of a literal $L$. Two
literals $L$ and $K$ are complementary if $\overline{L}=K$

From now on $D$ always denotes a finite ground clause set, also
called \it database, \rm and $\Sigma$ denotes its signature, i.e.,
the set of all predicate symbols occurring in it. We consider finite
ordered trees $T$ where the nodes, except the root node, are labeled
with literals. In the following we will represent a branch $b$ in
$T$ by the sequence $b=L_1,L_2,\ldots, L_n$ ($n\geq 0$) of its
literal labels, where $L_1$ labels an immediate successor of the
root node, and $L_n$ labels the leaf of $b$. The branch $b$ is
called \it regular \rm iff $L_i\neq L_j$ for $1\leq i,j\leq n$ and
$i\neq j$, otherwise it is called \it irregular. \rm The tree $T$ is
\it regular\rm~iff every of its branches is regular, otherwise it is
\it irregular. \rm The set of \it branch literals \rm of $b$ is
$lit(b)= \{L_1,L_2,\ldots,L_n\}$. For brevity, we will write
expressions like $A \in b$ instead of $A\in lit(b)$.  In order to
memorize the fact that a branch contains a contradiction, we allow
to label a branch as either \it open \rm or \it closed. \rm A
tableau is closed if each of its branches is closed, otherwise it is
open.

\begin{definition} [Hyper Tableau]
A literal set is called \it inconsistent \rm iff it contains a pair
of complementary literals, otherwise it is called \it consistent.
Hyper tableaux \rm for $D$ are inductively defined as follows: \bf

Initialization step: \rm  The empty tree, consisting of the root
node only, is a hyper tableau for $D$. Its single branch is marked
as "open". \bf

Hyper extension step: \rm  If (1) $T$ is an open hyper tableau for
$D$ with open branch $b$,  and (2) $C=A_1\lor A_2\lor\cdots\lor
A_m\leftarrow B_1\land B_2\cdots\land B_n$ is  a clause from $D$
($n\geq 0,m\geq 0$), called \it extending clause \rm in this
context, and (3) $\{B_1,B_2,\ldots, B_n\}\subseteq b$ (equivalently,
we say that $C$ is \it applicable to $b$)\rm  then the tree $T$ is a
hyper tableau for $D$, where $T$ is obtained from $T$ by extension
of $b$ by $C$:
 replace $b$ in $T$ by the \it new branches \rm \begin{equation*}
 (b,A_1),(b,A_2),\ldots,(b,A_m),(b,\neg B_1),(b,\neg B_2),\ldots,
 (b,\neg B_n)
 \end{equation*} and then mark every inconsistent new branch as "closed", and the
other new branches as "open".
\end{definition}

The applicability condition of an extension expresses that all body
literals have to be satisfied by the branch to be extended. From now
on, we consider only regular hyper tableaux. This restriction
guarantees that for finite clause sets no branch can be extended
infinitely often. Hence, in particular, no open finished branch can
be extended any further. This fact will be made use of below
occasionally. Notice as an immediate consequence of the above
definition that open branches never contain negative literals.

\subsection{View update algorithm}

The key idea of the algorithm presented in this paper is to
transform the given database along with the view update request into
a disjunctive logic program and apply known disjunctive techniques
to solve the original view update problem. The intuition behind the
transformation is to obtain a disjunctive logic program in such a
way that each (minimal) model of this transformed program represent
a way to update the given view atom. We present two variants of our
algorithm. The one that is discussed in this section employs a
trivial transformation procedure but has to look for minimal models;
and another performs a costly transformation, but dispenses with the
requirement of computing the minimal models.

\subsection{Minimality test}

We start presenting an algorithm for stratifiable (definite)
deductive databases by first defining precisely how the given
database is transformed into a disjunctive logic program for the
view deletion process \cite{Arav2} (successful branch - see in
Algorithms 3 and 4 via Hyper Tableau).

\begin{definition} [$IDB$ Transformation] Given an $IDB$ and a set of ground atoms $S$, the
transformation of $IDB$ wrt $S$ is obtained by translating each
clause $C\in IDB$ as follows: Every atom $A$ in the body (resp.
head) of $C$ that is also in $S$ is moved to the head (resp. body)
as $\neg A$.
\end{definition}

\begin{note} If $IDB$ is a stratifiable deductive database then the transformation
  introduced above is not necessary.
\end{note}

\begin{definition} [$IDB^*$ Transformation] Let $IDB\cup EDB$ be a given database.
Let $S_0=EDB\cup \{A~|~A~\text{ is a ground \it IDB \rm atom}\}$.
Then, $IDB^*$ is defined as the transformation of $IDB$ wrt $S_0$.
\end{definition}

\begin{note} Note that $IDB^*$ is in general a disjunctive logic program.
The negative literals $(\neg A)$ appearing in the clauses are
intuitively interpreted as deletion of the corresponding atom ($A$)
from the database. Technically, a literal $\neg A$ is to be read as
a \it positive \rm atom, by taking the $\neg$-sign as part of the
predicate symbol. To be more precise, we treat $\neg A$ as an atom
wrt $IDB^*$, but as a negative literal wrt $IDB$.

Note that there are no facts in $IDB^*$. So when we add a delete
request such as $\neg A$ to this, the added request is the only fact
and any bottom-up reasoning strategy is fully focused on the goal
(here the delete request)
\end{note}

\begin{definition}[Update Tableaux Hitting Set] An update tableau for a database
$IDB\cup EDB$ and delete request $\neg A$ is a hyper tableau $T$ for
$IDB^*\cup\{\neg A\leftarrow\}$ such that every open branch is
finished. For every open finished branch $b$ in $T$ we define the
\it hitting set (of b in $T$) \rm as $HS(b)=\{A \in EDB | \neg A \in
b\}$.
\end{definition}

\begin{definition}[Minimality test] Let $T$ be an update tableau for $IDB\cup
EDB$ and delete request $\neg A$. We say that open finished branch
$b$ in $T$ satisfies \it the strong minimality test \rm iff $\forall
s\in HS(b):IDB\cup EDB\backslash HS(b)\cup\{s\}\vdash A$.
\end{definition}

\begin{definition}[Update Tableau satisfying strong minimality] An update tableau
for given $IDB\cup  EDB$ and delete request $\neg A$ is transformed
into an update tableau satisfying strong minimality by marking every
open finished branch as closed which does not satisfy strong
minimality.
\end{definition}

Next step is view insertion process \cite{Beh} (For unsuccessful
branches - see in Algorithms 3 and 4 via magic set).

\begin{definition} [$IDB^**$ Transformation] Let $IDB\cup EDB$ be a given database.
Let $S_1=EDB\cup \{A~|~A~\text{ is a ground \it IDB \rm atom}\}$.
Then, $IDB^**$ is defined as the transformation of $IDB$ wrt $S_1$.
\end{definition}

\begin{note} Note that $IDB$ is in general a (stratifiable) disjunctive logic program.
The positive literals $(A)$ appearing in the clauses are intuitively
interpreted as an insertion of the corresponding atom ($A$) from the
database. \end{note}

\begin{definition}[Update magic Hitting Set] An update magic set rule for a database
$IDB\cup EDB$ and insertion request $A$ is a magic set rule $M$ for
$IDB^*\cup\{A\leftarrow\}$ such that every close branch is finished.
For every close finished branch $b$ in $M$ we define the \it magic
set rule (of b in $M$) \rm as $HS(b)=\{A \in EDB | A \in b\}$.
\end{definition}

\begin{definition}[Minimality test] Let $M$ be an update magic set rule for $IDB\cup
EDB$ and insert request $A$. We say that close finished branch $b$
in $M$ satisfies \it the strong minimality test \rm iff $\forall
s\in HS(b):IDB\cup EDB\backslash HS(b)\cup\{s\}\vdash \neg A$.
\end{definition}

\begin{definition}[Update magic set rule satisfying strong minimality] An update
magic set rule for given $IDB\cup  EDB$ and insert request $A$ is
transformed into an update magic set rule satisfying strong
minimality by marking every close finished branch as open which does
not satisfy strong minimality.
\end{definition}

This means that every minimal model (minimal wrt the base atoms) of
$IDB^* \cup \{\neg A\}$ provides a minimal hitting set for deleting
the ground view atom $A$. Similarly, $IDB^* \cup \{A\}$ provides a
minimal hitting set for inserting the ground view atom $A$. Now we
are in a position to formally present our algorithm. Given a
database and a view atom to be updated, we first transform the
database into a definite disjunctive logic program and use hyper
tableaux calculus to generate models of this transformed program for
deletion of an atom. Second, magic set rule is used to generate
models of this transformed program for insertion of an atom. Models
that do not represent rational update are filtered out using the
strong minimality test. This is formalized in Algorithm 3.

To show the rationality of this approach, we study how this is
related to the previous approach presented in the last section, i.e.
generating explanations and computing hitting sets of these
explanations. To better understand the relationship it is imperative
to study where the explanations are in the hyper tableau approach
and magic set rule. We first define the notion of $EDB$ -cut and
then view update seeds.

\begin{definition}[$EDB$-Cut] Let $T$ be update tableau with open branches
$b_1,b_2,\ldots,b_n$. A set $S=\{A_1,A_2,\ldots,A_n\}\subseteq EDB$
is said to be $EDB$-cut of $T$ iff  $\neg A_i\in b_i$ ($A_i\in
b_i$), for $1\leq i\leq n$.
\end{definition}

\begin{definition}[$EDB$ seeds]
Let $M$ be an update seeds with close branches $b_1,b_2,\ldots,b_n$.
A set $S=\{A_1,A_2,\ldots,A_n\}\subseteq EDB$ is said to be a
$EDB$-seeds of $M$ iff EDB seeds $vu\_seeds(\nu_D)$ with respect to
$\nu_D$ is defined as follows:
$$vu\_seeds(\nu_D) := \left\{\nabla^{\pi}_p (c_1,\ldots , c_n) | p(c_1,\ldots, c_n)\in \nu^{\pi}_D~and~\pi\in\{+,
-\}\right\} .$$ \rm
\end{definition}

$$\begin{array}{cc}\hline
\text{\bf Algorithm 3} &  \text{\rm View update algorithm based on
minimality test}\\\hline  \text{\rm
Input}:&
\text{\rm A stratifiable (definite) deductive database}~DDB=IDB\cup EDB\cup IC\\
&\text{\rm an literals}~\mathcal{A}\\
\text{\rm Output:}&\text{\rm A new stratifiable (definite) database}~IDB\cup EDB'\cup IC\\
\text{\rm begin}&\\
~~1.&\text{\rm Let}~ V :=\{ c\in IC~|~IDB\cup IC~\text{\rm
inconsistent
with}~\mathcal{A}~\text{\rm wrt}~c~\}\\
&\text{\rm While}~(V\neq \emptyset)\\
~~2.&\hspace{-1.9cm}\text{\rm Construct a complete SLD-tree for} \leftarrow\mathcal{A}~\text{\rm wrt DDB.}\\
~~3.&\hspace{-0.7cm}\text{\rm For every successful branch $i$:construct}~\Delta_{i}=\{D~|~D\in EDB \}\\
&\text{\rm and D is used as an input clause in branch $i$}.\\
&\text{\rm Construct a branch i of an update tableau satisfying
minimality}\\
&\text{\rm for}~ IDB\cup EDB~\text{\rm and delete request}~\neg A.\\
&\text{\rm  Produce}~IDB\cup EDB \backslash HS(i)~\text{\rm as a result}\\
~~4.&\text{\rm For every unsuccessful branch $j$:construct}~\Delta_{j}=\{D~|~D\in EDB \}\\
&\text{\rm and D is used as an input clause in branch $j$}.\\
&\text{\rm Construct a branch j of an update magic set rule
satisfying minimality}\\
&\text{\rm for}~ IDB\cup EDB~\text{\rm and insert request}~A.\\
&\text{\rm  Produce}~IDB\cup EDB \backslash HS(j)~\text{\rm as a
result}\\
&\text{\rm Let}~ V :=\{ c\in IC~|~IDB\cup IC~\text{\rm inconsistent
with}~\mathcal{A}~\text{\rm wrt}~c~\}\\
&\hspace{-0.7cm}\text{\rm return}\\
~~5.&\text{\rm Produce}~DDB~\text{\rm as the result.}\\
\text{\rm end.}&\\\hline
\end{array}$$

\begin{lemma} Let $T$ be an update tableau for $IDB\cup EDB$ and
update request $A$. Similarly, for $M$ be an update magic set rule.
Let $S$ be the set of all $EDB$-closed minimal abductive
explanations for $A$ wrt. $IDB$.  Let $S'$ be the set of all
$EDB$-cuts of $T$ and $EDB$-seeds of $M$ . Then the following hold
\begin{enumerate}
\item[$\bullet$] $S\subseteq S'$.\\
\item[$\bullet$] $\forall \Delta'\in S':\exists \Delta\in S s.t. \Delta\subseteq \Delta'$.
\end{enumerate}
\end{lemma}

The above lemma precisely characterizes what explanations are
generated by an update tableau. It is obvious then that a branch
cuts through all the explanations and constitutes a hitting set for
all the generated explanations. This is formalized below.

\begin{lemma} Let $S$ and $S'$ be sets of sets s.t. $S\subseteq S'$ and
every member  of $S'\backslash S$ contains an element of S. Then, a
set $H$ is a minimal hitting set for $S$ iff it is a minimal hitting
set for $S'$.
\end{lemma}

\begin{lemma}Let $T$ be an update tableau for $IDB\cup EDB$ and
update request $A$ that satisfies the strong minimality test.
Similarly, for $M$ be an update magic set rule. Then, for every open
(close) finished branch $b$ in $T$, $HS(b)$ ($M$, $HS(b)$) is a
minimal hitting set of all the abductive explanations of $A$.
\end{lemma}

So, Algorithms 3 generate a minimal hitting set (in polynomial
space) of all $EDB$-closed locally minimal abductive explanations of
the view atom to be deleted. From the belief dynamics results
recalled in section 3, it immediately follows that Algorithms 5 and
6 are rational, and satisfy the strong relevance postulate (KB-7.1).

\begin{theorem} Algorithms 3 is a rational, in the sense that
they satisfy all the rationality postulates (KB*1)-(KB*6) and the
strong relevance postulate (KB*7.1). Further, any update that
satisfies these postulates can be computed by these algorithms.
\end{theorem}

\subsection{Materialized view}
In many cases, the view to be updates is materialized, i.e., the
least Herbrand Model is computed and kept, for efficient query
answering. In such a situation, rational hitting sets can be
computed without performing any minimality test. The idea is to
transform the given $IDB$ wrt the materialized view.

\begin{definition} [$IDB^+$ Transformation] Let $IDB\cup  EDB$ be a given
database. Let $S$ be the Least Herbrand Model of this database.
Then, $IDB^+$ is defined as the transformation of $IDB$ wrt $S$.
\end{definition}

\begin{note}
If $IDB$ is a stratifiable deductive database then the
transformation introduced above is not necessary.
\end{note}

\begin{definition}[Update Tableau based on Materialized view] An update tableau
based on materialized view for a database $IDB\cup EDB$ and delete
request $\neg A$ is a hyper tableau $T$ for $IDB^+\cup\{\neg
A\leftarrow \}$ such that  every open branch is finished.
\end{definition}

\begin{definition} [$IDB^-$ Transformation] Let $IDB\cup  EDB$ be a given
database. Let $S_1$ be the Least Herbrand Model of this database.
Then, $IDB^-$ is defined as the transformation of $IDB$ wrt $S_1$.
\end{definition}

\begin{definition}[Update magic set rule based on Materialized view] An update
magic set rule based on materialized view for a database $IDB\cup
EDB$ and insert request $A$ is a magic set $M$ for
$IDB^+\cup\{A\leftarrow \}$ such that  every close branch is
finished.
\end{definition}

Now the claim is that every model of $IDB^+\cup\{\neg A\leftarrow
\}$ ($A\leftarrow$) constitutes a rational hitting set for the
deletion and insertion of the ground view atom $A$. So, the
algorithm works as follows: Given a database and a view update
request, we first transform the database wrt its Least Herbrand
Model (computation of the Least Herbrand Model can be done as a
offline preprocessing step. Note that it serves as materialized view
for efficient query answering). Then the hyper tableaux calculus
(magic set rule) is used to compute models of this transformed
program. Each model represents a rational way of accomplishing the
given view update request. This is formalized in Algorithms 4.

This approach for view update may not satisfy (KB*7.1) in general.
But, as shown in the sequel, conformation to(KB*6.3) is guaranteed
and thus this approach results in rational update.

$$\begin{array}{cc}\hline
\text{\bf Algorithm 4} &  \text{\rm View update algorithm based on
Materialized view}\\\hline\text{\rm
Input}:&
\text{\rm A stratifiable (definite) deductive database}~DDB=IDB\cup EDB\cup IC\\
&\text{\rm an literals}~\mathcal{A}\\
\text{\rm Output:}&\text{\rm A new stratifiable (definite) database}~IDB\cup EDB'\cup IC\\
\text{\rm begin}&\\
~~1.&\text{\rm Let}~ V :=\{ c\in IC~|~IDB\cup IC~\text{\rm
inconsistent
with}~\mathcal{A}~\text{\rm wrt}~c~\}\\
&\text{\rm While}~(V\neq \emptyset)\\
~~2.&\hspace{-1.9cm}\text{\rm Construct a complete SLD-tree for} \leftarrow\mathcal{A}~\text{\rm wrt DDB.}\\
~~3.&\hspace{-0.7cm}\text{\rm For every successful branch $i$:construct}~\Delta_{i}=\{D~|~D\in EDB \}\\
&\text{\rm and D is used as an input clause in branch $i$}.\\
&\text{\rm Construct a branch i of an update tableau based on view}\\
&\text{\rm for}~ IDB\cup EDB~\text{\rm and delete request}~\neg A.\\
&\text{\rm  Produce}~IDB\cup EDB \backslash HS(i)~\text{\rm as a result}\\
~~4.&\text{\rm For every unsuccessful branch $j$:construct}~\Delta_{j}=\{D~|~D\in EDB \}\\
&\text{\rm and D is used as an input clause in branch $j$}.\\
&\text{\rm Construct a branch j of an update magic set rule
based on view}\\
&\text{\rm for}~ IDB\cup EDB~\text{\rm and insert request}~A.\\
&\text{\rm  Produce}~IDB\cup EDB \backslash HS(j)~\text{\rm as a
result}\\
&\text{\rm Let}~ V :=\{ c\in IC~|~IDB\cup IC~\text{\rm inconsistent
with}~\mathcal{A}~\text{\rm wrt}~c~\}\\
&\hspace{-0.7cm}\text{\rm return}\\
~~5.&\text{\rm Produce}~DDB~\text{\rm as the result.}\\
\text{\rm end.}&\\\hline
\end{array}$$

\begin{lemma} Let $T$ be an update tableau based on materialized view for
$IDB\cup  EDB$ and delete request $\neg A$ ($A$), Similarly, for $M$
be an update magic set rule. Let $S$ be the set of all $EDB$-closed
locally minimal abductive explanations for $A$ wrt $IDB$. Let $S'$
be the set of all $EDB$-cuts of $T$ and  $EDB$-seeds of $M$. Then,
the following hold:
\begin{enumerate}
\item[$\bullet$] $S\subseteq S'$.
\item[$\bullet$] $\forall \Delta'\in S':\exists \Delta\in S ~ s.t.~ \Delta\subseteq \Delta'$.
\item[$\bullet$] $\forall \Delta'\in S':\Delta'\subseteq \bigcup S$.
\end{enumerate}
\end{lemma}

\begin{lemma} Let $S$ and $S'$ be sets of sets s.t. $S\in  S'$ and for every
member $X$ of $S'\backslash S$: $X$ contains a member of $S$ and $X$
is contained in $\bigcup S$. Then, a set $H$ is a hitting set for
$S$ iff it is a hitting set for $S'$.
\end{lemma}

\begin{lemma} Let $T$ and $M$ be defined as in Lemma 5. Then $HS(b)$ is a rational hitting set
for $A$, for every open finished branch $b$ in $T$ (close finished
branch $b$ in $M$).
\end{lemma}

\begin{theorem} Algorithms 4 is a rational, in the sense that they satisfy all
the rationality postulates (KB*1) to (KB*6) and (KB*7.3).
\end{theorem}

\section{A Comparative Study of Integrity Constraints and View Update}
During the process of updating a database, two interrelated problems
could arise. On one hand, when an update is applied to the database,
integrity constraints could become inconsistent with request, then
stop the process. On the other hand, when an update request consist
on updating some derived predicate, a view updating mechanism must
be applied to translate the update request into correct updates on
the underlying base facts. Our work is not focusing on the integrity
constraint maintenance approach. In this section, we extend Mayol
and Teniente's \cite{Mayol} survey for view updating and integrity
constraint.

The main aspects that must be taken into account during the process
of view updating and integrity constraint \cite{Fra} enforcement are the
following: the problem addressed, the considered database schema,
the allowed update requests, the used technique, update change and
the obtained solutions. These six aspects provide the basic
dimensions to be taken into account. We explain each dimension in
this section.\\

\textbf{Problem Addressed}
\begin{enumerate}
\item[]\hspace{-0.6cm}(\emph{Type})~-~What kind of program to be used (stratified(S), Horn clause(H), Disjunctive database(D), Normal Logic program(N) and Other (O)).
\item[]\hspace{-0.6cm}(\emph{View Update})~-~Whether they are able to deal with view updating or not (indicated by Yes or
No in the second column of Table 1).
\item[]\hspace{-0.6cm}(\emph{integrity-constraint Enforcement})~-~Whether they incorporate an integrity constraint checking or an integrity
constraint maintenance approach (indicated by check or maintain in
the third column).
\item[]\hspace{-0.6cm}(\emph{Run/Comp})~-~Whether the method follows a run-time(transaction) or a compile-time approach (indicated
by Run or Compile in the fourth column).
\end{enumerate}

\textbf{Database Schema Considered}
\begin{enumerate}
\item[]\hspace{-0.6cm}(\emph{Definition Language})~-~The language mostly used is logic, although some
methods use a relational language and also uses an object-oriented.
\item[]\hspace{-0.6cm}(\emph{The DB Schema Contains Views})~-~ All methods
that deal with view updating need views to be defined in the
database schema. Some of other method allow to define views.
\item[]\hspace{-0.6cm}(\emph{Restrictions Imposed on the Integrity Constraints})~-~ Some proposals impose
certain restrictions on the kind of integrity constraints that can
be defined and, thus, handled by their methods.
\item[]\hspace{-0.6cm}(\emph{Static vs Dynamic Integrity Constraints})~-~ Integrity
constraints may be either static, and impose restrictions involving
only a certain state of the database, or dynamic.
\end{enumerate}

\textbf{Update Request Allowed}
\begin{enumerate}
\item[]\hspace{-0.6cm}(\emph{Multiple Update Request})~-~An update request is multiple if it contains several
updates to be applied together to the database.
\item[]\hspace{-0.6cm}(\emph{Update Operators})~-~Traditionally, three different basic update operators are
distinguished: insertion ($\iota$), deletion ($\delta$) and
modification ($\chi$). Modification can always be simulated by a
deletion followed by an insertion.
\end{enumerate}

\textbf{Update Processing Mechanism}
\begin{enumerate}
\item[]\hspace{-0.6cm}(\emph{Applied Technique})~-~ The techniques applied by these methods can be classified
according to four different kinds of procedures, unfolding, SLD,
active and predefined programs, respectively.
\item[]\hspace{-0.6cm}(\emph{Taking Base Facts into Account}~-~ Base facts can either be taken into
account or not during update processing.
\item[]\hspace{-0.6cm}(\emph{User Participation})~-~ User participation during update
processing or not.
\end{enumerate}

\textbf{Update Changing Mechanism}
\begin{enumerate}

\item[]\hspace{-0.6cm}(\emph{Type of modification})~-~
Changing table by singleton like atom (S), sets of each types of
modification(SS) and group of changes(G).
\item[]\hspace{-0.6cm}(\emph{Changing Base Fact})~-~
Base fact can be changed either using principle of minimal change or
complete change (maximal change).
\item[]\hspace{-0.6cm}(\emph{Changing View Definition})~-~ Whether
update process view definition is changed or not.
\end{enumerate}

\textbf{Obtained Solution}
\begin{enumerate}
\item[]\hspace{-0.6cm}(\emph{Our Axiom follow})~-~ When update
process done, we are comparing our axiomatized method and which
relevance policy holds ((KB*1) to (KB*6),(KB*7.1),(KB*7.2) and
(KB*7.3) is enumerated 1 to 9)
\item[]\hspace{-0.6cm}(\emph{Soundness})~-~ A
method is correct if it only obtains solutions that satisfy the
requested update.
\item[]\hspace{-0.6cm}(\emph{Completeness)}~-~ A method is complete if it is able to obtain all
solutions that satisfy a given update request.
\end{enumerate}

Results of each method according to these features are summarized in
Appendix Table 1.


\section{Related Works}

We begin by recalling previous work on view deletion. Chandrabose
\cite{Arav1,Arav} and Delhibabu \cite{Del,Del1,Del2}, defines a
contraction and revision operator in view deletion with respect to a
set of formulae or sentences using Hansson's \cite{Hans2} belief
change. Similar to our approach, he focused on set of formulae or
sentences in knowledge base revision for view update wrt. insertion
and deletion and formulae are considered at the same level.
Chandrabose proposed different ways to change knowledge base via
only database deletion, devising particular postulate which is shown
to be necessary and sufficient for such an update process.

Our Horn knowledge base consists of two parts, immutable part and
updatable part , but focus is on principle of minimal change. There
are more related works on that topic. Eiter \cite{Eit},
Langlois\cite{Lang}, and Delgrande \cite{Delg} are focusing on Horn
revision with different perspectives like prime implication, logical
closure and belief level. Segerberg \cite{Seg} defined new modeling
for belief revision in terms of irrevocability on prioritized
revision. Hansson \cite{Hans2}, constructed five types of
non-prioritized belief revision. Makinson \cite{Mak} developed
dialogue form of revision AGM. Papini\cite{Pap} defined a new
version of knowledge base revision. Here, we consider immutable part
as a Horn clause and updatable part as an atom(literals).

We are bridging gap between philosophical work, paying little
attention to computational aspects of database work. In such a case,
Hansson's\cite{Hans2} kernel change is related with abductive
method. Aliseda's \cite{Alis} book on abductive reasoning is one of
the motivation keys. Christiansen's \cite{Chris} work on dynamics of
abductive logic grammars exactly fits our minimal change (insertion
and deletion). Wrobel's \cite{Wrob} definition of  first order
theory revision was helpful to frame our algorithm.

On other hand, we are dealing with view update problem. Keller's
\cite{Kell} thesis is motivation  for view update problem. There is
a lot of papers on view update problem (for example, recent survey
paper on view update by Chen and Liao\cite{Chen}, survey paper on
view algorithm by Mayol and Teniente \cite{Mayol} and current survey
paper on view selection (\cite{Dho,Hal,Li,Lee,Mam,Zhan}).
More similar to our work is paper presented by Bessant et al.
\cite{Bess} , local search-based heuristic technique that
empirically proves to be often viable, even in the context of very
large propositional applications. Laurent et al.\cite{Lau} parented
updating deductive databases in which every insertion or deletion of
a fact can be performed in a deterministic way.

Furthermore, and at a first sight more related to our work, some
work has been done on ontology systems and description logics (Qi
and Yang \cite{Qi}, and Kogalovsky \cite{Kog}). Finally, when we presented connection between
belief update versus database update, we did not talk about
complexity (see the works of Liberatore \cite{Lib1,Lib2}, Caroprese
\cite{Caro}, Calvanese's \cite{Cal}, and Cong \cite{Cong}).

The significance of our work can be summarized in the following:
\begin{description}
  \item[-] We have defined new way of insertion and deletion of an atom(literals) as per
  norm of principle of minimal change.
  \item[-] We have proposed new generalized revision algorithm for
  knowledge base dynamics, interesting connections with kernel change and
  abduction procedure.
   \item[-] We have written new view update algorithm for DDB, and
  we provided stratifiable (definite) deductive database, using our axiomatic method based on Hyper tableaux
  and magic sets.
  \item[-] Finally, we presented current Comparative Study of view update algorithms.

\end{description}


\section{Conclusion and remarks}

The main contribution of this research is to provide a link between
theory of belief dynamics and concrete applications such as view
updates in databases. We argued for generalization of belief
dynamics theory in two respects: to handle certain part of knowledge
as immutable; and dropping the requirement that belief state be
deductively closed. The intended generalization was achieved by
introducing the concept of knowledge base dynamics and generalized
contraction for the same. Further, we also studied the relationship
between  knowledge base dynamics and abduction resulting in a
generalized algorithm for revision based on abductive procedures. We
also successfully demonstrated how knowledge base dynamics can
provide an axiomatic characterization for updating an atom(literals)
to a stratifiable (definite) deductive database.

In bridging the gap between belief dynamics and view updates, we
have observed that a balance has to be achieved between
computational efficiency and rationality. While rationally
attractive notions of generalized revision prove to be
computationally inefficient, the rationality behind efficient
algorithms based on incomplete trees is not clear at all. From the
belief dynamics point of view, we may have to sacrifice some
postulates, vacuity for example, to gain computational efficiency.
Further weakening of relevance has to be explored, to provide
declarative semantics for algorithms based on incomplete trees.

On the other hand, from the database side, we should explore various
ways of optimizing the algorithms that would comply with the
proposed declarative semantics. We believe that partial deduction
and loop detection techniques, will play an important role in
optimizing algorithms of the previous section. Note that, loop
detection could be carried out during partial deduction, and
complete SLD-trees can be effectively constructed wrt a partial
deduction (with loop check) of a database, rather than wrt database
itself. Moreover, we would anyway need a partial deduction for
optimization of query evaluation.

We have presented two variants of an algorithm for updating a view
atom to a definite database. The key idea of this approach is to
transform the given database into a disjunctive logic program in
such a way that updates can be read off from the models of this
transformed program. One variant based on materialized views is of
polynomial time complexity. Moreover, we have also shown that this
algorithm is rational in the sense that it satisfies the rationality
postulates that are justified from philosophical angle.

In the second variant, where materialized view is used for the
transformation, after generating a hitting set and removing
corresponding $EDB$ atoms, we easily move to the new materialized
view. An obvious way is to recompute the view from scratch using the
new $EDB$ (i.e. compute the Least Herbrand Model of the new updated
database from scratch) but it is certainly interesting to look for
more efficient methods. In the end, we plan to redefined the model
to Horn Logic with stratified Negation \cite{Jack} and Argumentative
Inference \cite{Fala,Hera}.

Though we have discussed only about view updates, we believe that
knowledge base dynamics can also be applied to other applications
such as view maintenance, diagnosis, and we plan to explore it
further (see works \cite{Caro} and \cite{Bis}). It would also be
interesting to study how results using soft stratification
\cite{Beh} with belief dynamics, especially the relational approach,
could be applied in real world problems. Still, a lot of
developments are possible, for improving existing operators or for
defining new classes of change operators. As immediate extension,
question raises: is there any \emph{real life application for AGM in
25 year theory?} \cite{Ferme}. The revision and update are more
challenging in logical view update problem(database theory), so we
can extend the theory to combine results similar to Konieczny's
\cite{Kon} and Nayak's \cite{Nayak2}.


\section*{Appendix}

\text{Proof of Theorem 1}. Sound and Completeness are trivially to
shown from the definition. $\blacksquare$

\vspace{0.1cm}

\text{Proof of Lemma 1}.
\begin{enumerate}
\item[1.] \rm The fact that $\nvdash\neg\alpha$ and there exists a KB -
closed locally minimal abductive explanation for $\alpha$ wrt
$KB_{I}$, it is clear that there exists at least one $\alpha$-
kernel of KB. Suppose $\Delta$ ($\Delta\in\Delta^{+} \cup
\Delta^{-}$) is empty (i.e., $KB_{I}\vdash\neg\alpha$), then the
required result follows immediately. If not, since $\Delta$ is a
locally minimal abductive explanation, there exists a minimal subset
$KB_{I}'\subseteq KB_{I}$, s.t. $\Delta$ is minimal abductive
explanation of $\alpha$ wrt $KB_{I}'$. Since, $\Delta$ is KB-closed,
it is not difficult to see that $KB_{I}'\cup \Delta^{+} \cup
\Delta^{-}$ is a $\alpha$ - kernel of KB.
\item[2.] Since X is a $\alpha$ - kernel of KB and $\Delta$ is the
set of all abducibles in X, it follows that $\Delta^{+} \cup
\Delta^{-}$ is a minimal abductive explanation of $\Delta$ wrt
$X\backslash \Delta^{-} \cup \Delta^{+}$. It is obvious that
$\Delta^{+} \cup \Delta^{-}$ is KB- closed, and so $\Delta$ is a
KB-closed locally minimal abductive explanation for $\alpha$ wrt
$KB_{I}$. $\blacksquare$
\end{enumerate}

\vspace{0.1cm}

\text{Proof of Theorem 2}. Sound and Completeness are trivially to
shown from the Algorithm 1. $\blacksquare$

\text{Proof of Lemma 2 and 5}.
\begin{enumerate}
\item[1.] Consider a $\Delta (\Delta\in\Delta_i \cup \Delta_j)\in S$.
 We need to show that $\Delta$ is generated by algorithm
3 at step 2. From lemma 1, it is clear that there exists a
$A$-kernel $X$ of $DDB_G$ s.t. $X \cap EDB = \Delta_j$ and $X \cup
EDB = \Delta_i$. Since $X \vdash A$, there must exist a successful
derivation for $A$ using only the elements of $X$ as input clauses
and similarly $X \nvdash A$. Consequently $\Delta   $ must have been
constructed at step 2.
\item[2.] Consider a $\Delta'((\Delta'\in\Delta_i \cup \Delta_j)\in S'$. Let $\Delta'$ be
constructed from a successful(unsuccessful) branch $i$ via
$\Delta_i$($\Delta_j$). Let $X$ be the set of all input clauses used
in the refutation $i$. Clearly $X\vdash A$($X\nvdash A$). Further,
there exists a minimal (wrt set-inclusion) subset $Y$ of $X$ that
derives $A$ (i.e. no proper subset of $Y$ derives $A$). Let $\Delta
= Y \cap EDB$ ($Y \cup EDB$). Since IDB does not(does) have any unit
clauses, $Y$ must contain some EDB facts, and so $\Delta$ is not
empty (empty) and obviously $\Delta\subseteq \Delta'$. But, $Y$ need
not (need) be a $A$-kernel for $IDB_G$ since $Y$ is not ground in
general. But it stands for several $A$-kernels with the same
(different) EDB facts $\Delta$ in them. Thus, from lemma 1, $\Delta$
is a DDB-closed locally minimal abductive explanation for $A$ wrt
$IDB_G$ and is contained in $\Delta'$.
\item[3.] Since this proof easy to see materialized view update with
minimal.
\end{enumerate}

\vspace{0.1cm}

\text{Proof of Lemma 3 and 6}.
\begin{enumerate}
\item[1.] (\textbf{Only if part})~Suppose $H$ is a minimal hitting set for $S$. Since $S
\subseteq S'$ , it follows that $H \subseteq \bigcup S'$ . Further,
$H$ hits every element of $S'$ , which is evident from the fact that
every element of $S'$ contains an element of $S$. Hence $H$ is a
hitting set for $S'$ . By the same arguments, it is not difficult to
see that $H$ is minimal for $S'$ too.\\

(\textbf{If part})~Given that $H$ is a minimal hitting set for $S'$
, we have to show that it is a minimal hitting set for $S$ too.
Assume that there is an element $E \in H$ that is not in $\bigcup
S$. This means that $E$ is selected from some $Y \in S'\backslash
S$. But $Y$ contains an element of $S$, say $X$. Since $X$ is also a
member of $S'$ , one member of $X$ must appear in $H$. This implies
that two elements have been selected from $Y$ and hence $H$ is not
minimal. This is a contradiction and hence $H \subseteq \bigcup S$.
Since $S \subseteq S'$ , it is clear that $H$ hits every element in
$S$, and so $H$ is a hitting set for $S$. It remains to be shown
that $H$ is minimal. Assume the contrary, that a proper subset $H'$
of $H$ is a hitting set for $S$. Then from the proof of the only if
part, it follows that $H'$ is a hitting set for $S'$ too, and
contradicts the fact that $H$ is a minimal hitting set for $S'$ .
Hence, $H$ must be a minimal hitting set for $S$.\\

\item[2.] (\textbf{If part})~Given that $H$ is a hitting set for $S'$ , we have to
show that it is a hitting set for $S$ too. First of all, observe
that $\bigcup S = \bigcup S'$ , and so $H \subseteq \bigcup S$.
Moreover, by definition, for every non-empty member $X$ of $S'$ , $H
\cap X$ is not empty. Since $S \subseteq S'$ , it follows that $H$
is a hitting set for $S$ too.\\

(\textbf{Only if part})~Suppose $H$ is a hitting set for $S$. As
observed above, $H \subseteq \bigcup S'$ . By definition, for every
non-empty member $X\in S$, $X \cap H$ is not empty. Since every
member of $S'$ contains a member of $S$, it is clear that $H$ hits
every member of $S'$ , and hence a hitting set for $S'$ .
$\blacksquare$
\end{enumerate}

\vspace{0.1cm}

\text{Proof of Lemma 4 and 7}. Follows from the lemma 3,4 (minimal
test) and 6,7 (materialized view) of \cite{Beh} $\blacksquare$

\vspace{0.1cm}

 \text{Proof of Theorem 3}. Follows from Lemma 3 and Theorem 2. $\blacksquare$

\vspace{0.1cm}

\text{Proof of Theorem 4}. Follows from Lemma 6 and Theorem 2.
$\blacksquare$

\begin{sidewaystable}
\centering
\begin{turn}{180}
\begin{tabular}{|l|l|l|l|l|l|l|l|l|l|l|l|l|l|l|l|l|l|l|l|}\hline
\multirow{3}{*}{\bf Method}&\multicolumn{4}{c}{\bf Problem}&
\multicolumn{4}{|c|}{\bf Database
  schema}&\multicolumn{2}{c}{\bf Update req.}&\multicolumn{3}{|c|}{\bf Mechanism}&\multicolumn{3}{c}{\bf Update Change}&
\multicolumn{3}{|c|}{\bf Solutions}\\
\cline{2-20}

 &\multirow{2}{*}{Type}&View&IC&Run/&Def.&\multirow{2}{*}{View}&IC&Kind of
&\multirow{2}{*} {Mul.}&Update&Tech-&Base&User&\multirow{2}{*}{Type}&~Base~&View&\multirow{2}{*}{Axiom}&\multirow{2}{*}{Sound.}&\multirow{2}{*}{Complete.}\\
&&Update&Enforce.&Comp.&Lang.&&def.&IC&&Operat.&nique&Facts&Part.&&facts&def.&&&\\\hline

\multirow{2}{*}{\bf
\cite{Gue}}&\multirow{2}{*}{N}&\multirow{2}{*}{Yes}&\multirow{2}{*}{Check}&\multirow{2}{*}{Run}
&\multirow{2}{*}{Logic}&\multirow{2}{*}{Yes}&\multirow{2}{*}{Yes}&\multirow{2}{*}{Static}&\multirow{2}{*}{No}&\multirow{2}{*}{
$\iota~\delta$}&\multirow{2}{*}{
SLDNF}&\multirow{2}{*}{No}&\multirow{2}{*}{No}&\multirow{2}{*}{S}&\multirow{2}{*}{Yes}&\multirow{2}{*}{No}&\multirow{2}{*}{1-6,9}&\multirow{2}{*}{No}&Not\\
&&&&&&&&&&&&&&&&&&&proved\\\hline

\multirow{2}{*}{\bf
\cite{Kak}}&\multirow{2}{*}{N}&\multirow{2}{*}{Yes}&\multirow{2}{*}{Maintain}&\multirow{2}{*}{Run}
&\multirow{2}{*}{Logic}&\multirow{2}{*}{Yes}&\multirow{2}{*}{Yes}&\multirow{2}{*}{Static}&\multirow{2}{*}{Yes}&\multirow{2}{*}{
$\iota~\delta$}&\multirow{2}{*}{
SLDNF}&\multirow{2}{*}{No}&\multirow{2}{*}{No}&\multirow{2}{*}{S}&\multirow{2}{*}{Yes}&\multirow{2}{*}{No}&\multirow{2}{*}{---}&\multirow{2}{*}{No}&\multirow{2}{*}{No}\\
&&&&&&&&&&&&&&&&&&&\\\hline

\multirow{2}{*}{\bf
\cite{Kuc}}&\multirow{2}{*}{S}&\multirow{2}{*}{Yes}&\multirow{2}{*}{Check}&\multirow{2}{*}{Run}
&\multirow{2}{*}{Logic}&\multirow{2}{*}{Yes}&\multirow{2}{*}{Yes}&\multirow{2}{*}{Static}&\multirow{2}{*}{Yes}&\multirow{2}{*}{
$\iota~\delta$}&\multirow{2}{*}{
---}&\multirow{2}{*}{Yes}&\multirow{2}{*}{No}&\multirow{2}{*}{SS}&\multirow{2}{*}{Yes}&\multirow{2}{*}{Yes}&\multirow{2}{*}{1-6,7}&Not&Not\\
&&&&&&&&&&&&&&&&&&proved&proved\\\hline

\multirow{2}{*}{\bf
\cite{Moer}}&\multirow{2}{*}{N}&\multirow{2}{*}{No}&\multirow{2}{*}{Maintain}&\multirow{2}{*}{Run}
&\multirow{2}{*}{Logic}&\multirow{2}{*}{Yes}&\multirow{2}{*}{Yes}&\multirow{2}{*}{Static}&\multirow{2}{*}{Yes}&\multirow{2}{*}{
$\iota~\delta$}&\multirow{2}{*}{
---}&\multirow{2}{*}{Yes}&\multirow{2}{*}{No}&\multirow{2}{*}{S}&\multirow{2}{*}{Yes}&\multirow{2}{*}{No}&\multirow{2}{*}{1-6,7}
&No&No\\
&&&&&&&&&&&&&&&&&&proved&proved\\\hline

\multirow{2}{*}{\bf
\cite{Gup}}&\multirow{2}{*}{S}&\multirow{2}{*}{Yes}&Check&Comp.
&Relation.&\multirow{2}{*}{No}&\multirow{2}{*}{No}&\multirow{2}{*}{Static}&\multirow{2}{*}{Yes}&\multirow{2}{*}{
$\iota~\delta~\chi$}&predef.&\multirow{2}{*}{Yes}&\multirow{2}{*}{No}&\multirow{2}{*}{G}&\multirow{2}{*}{Yes}&\multirow{2}{*}{No}&\multirow{2}{*}{---}
&\multirow{2}{*}{No}&\multirow{2}{*}{No}\\
&&&Maintain&Run&Logic&&&&&&Programs&&&&&&&&\\\hline

\multirow{2}{*}{\bf
\cite{Don}}&\multirow{2}{*}{N}&\multirow{2}{*}{Yes}&\multirow{2}{*}{Check}&\multirow{2}{*}{Run}
&\multirow{2}{*}{Logic}&\multirow{2}{*}{Yes}&\multirow{2}{*}{No}&\multirow{2}{*}{Static}&\multirow{2}{*}{Yes}&\multirow{2}{*}{
$\iota~\delta$}&predef&\multirow{2}{*}{Yes}&\multirow{2}{*}{No}&\multirow{2}{*}{S}&\multirow{2}{*}{Yes}&\multirow{2}{*}{No}&\multirow{2}{*}{1-6,7}&Not&\multirow{2}{*}{No}\\
&&&&&&&&&&&Programs&&&&&&&Proved&\\\hline

\multirow{2}{*}{\bf
\cite{Urp}}&\multirow{2}{*}{S}&\multirow{2}{*}{Yes}&Check&\multirow{2}{*}{Run}
&\multirow{2}{*}{Logic}&\multirow{2}{*}{Yes}&\multirow{2}{*}{Yes}&\multirow{2}{*}{Static}&\multirow{2}{*}{Yes}&\multirow{2}{*}{
$\iota~\delta~\chi$}&\multirow{2}{*}{
SLDNF}&\multirow{2}{*}{No}&\multirow{2}{*}{No}&\multirow{2}{*}{SS}&\multirow{2}{*}{Yes}&\multirow{2}{*}{No}&\multirow{2}{*}{1-6,7}&\multirow{2}{*}{Yes}&\multirow{2}{*}{No}\\
&&&Maintain&&&&&&&&&&&&&&&&\\\hline

\multirow{2}{*}{\bf
\cite{Gup1}}&\multirow{2}{*}{N}&\multirow{2}{*}{Yes}&\multirow{2}{*}{Maintain}&\multirow{2}{*}{Run}
&\multirow{2}{*}{Logic}&\multirow{2}{*}{Yes}&\multirow{2}{*}{No}&\multirow{2}{*}{Static}&\multirow{2}{*}{Yes}&\multirow{2}{*}{
$\iota~\delta$}&\multirow{2}{*}{Unfold}&\multirow{2}{*}{Yes}&\multirow{2}{*}{No}&\multirow{2}{*}{SS}&\multirow{2}{*}{Yes}&\multirow{2}{*}{No}&\multirow{2}{*}{1-6,9}&\multirow{2}{*}{Yes}&\multirow{2}{*}{Yes}\\
&&&&&&&&&&&&&&&&&&&\\\hline

\multirow{2}{*}{\bf
\cite{May}}&\multirow{2}{*}{N}&\multirow{2}{*}{Yes}&\multirow{2}{*}{Maintain}&Comp.
&\multirow{2}{*}{Logic}&\multirow{2}{*}{Yes}&\multirow{2}{*}{Yes}&Static&\multirow{2}{*}{Yes}&\multirow{2}{*}{
$\iota~\delta~\chi$}&\multirow{2}{*}{
SLDNF}&\multirow{2}{*}{Yes}&\multirow{2}{*}{No}&\multirow{2}{*}{S}&\multirow{2}{*}{Yes}&\multirow{2}{*}{No}&\multirow{2}{*}{1-6,9}&Not&Not\\
&&&&Run&&&&Dynamic&&&&&&&&&&proved&proved\\\hline

\multirow{2}{*}{\bf
\cite{Wut}}&\multirow{2}{*}{S}&\multirow{2}{*}{Yes}&\multirow{2}{*}{Maintain}&\multirow{2}{*}{Run}
&\multirow{2}{*}{Logic}&\multirow{2}{*}{Yes}&\multirow{2}{*}{Yes}&\multirow{2}{*}{Static}&\multirow{2}{*}{Yes}&\multirow{2}{*}{
$\iota~\delta$}&\multirow{2}{*}{
Unfold.}&\multirow{2}{*}{No}&\multirow{2}{*}{Yes}&\multirow{2}{*}{S}&\multirow{2}{*}{Yes}&\multirow{2}{*}{No}&\multirow{2}{*}{1-6,7}&Not&\multirow{2}{*}{No}\\
&&&&&&&&&&&&&&&&&&proved&\\\hline

\multirow{2}{*}{\bf
\cite{Arav1}}&\multirow{2}{*}{H}&\multirow{2}{*}{Yes}&\multirow{2}{*}{Check}&\multirow{2}{*}{Run}
&\multirow{2}{*}{Logic}&\multirow{2}{*}{Yes}&\multirow{2}{*}{Yes}&\multirow{2}{*}{Static}&\multirow{2}{*}{Yes}&\multirow{2}{*}{
$\iota~\delta$}&\multirow{2}{*}{
SLD}&\multirow{2}{*}{Yes}&\multirow{2}{*}{No}&\multirow{2}{*}{S}&\multirow{2}{*}{Yes}&\multirow{2}{*}{No}&\multirow{2}{*}{1-6,9}&\multirow{2}{*}{Yes}&\multirow{2}{*}{Yes}\\
&&&&&&&&&&&&&&&&&&&\\\hline

\multirow{2}{*}{\bf
\cite{Cer}}&\multirow{2}{*}{N}&\multirow{2}{*}{No}&\multirow{2}{*}{Maintain}&Comp
&Relation&\multirow{2}{*}{Yes}&\multirow{2}{*}{Limited}&\multirow{2}{*}{Static}&\multirow{2}{*}{Yes}&\multirow{2}{*}{
$\iota~\delta~\chi$}&\multirow{2}{*}{
Active}&\multirow{2}{*}{Yes}&\multirow{2}{*}{Yes}&\multirow{2}{*}{S}&\multirow{2}{*}{Yes}&\multirow{2}{*}{No}&\multirow{2}{*}{---}&\multirow{2}{*}{No}&\multirow{2}{*}{No}\\
&&&&Run&Logic&&&&&&&&&&&&&&\\\hline

\multirow{2}{*}{\bf
\cite{Ger}}&\multirow{2}{*}{N}&\multirow{2}{*}{No}&\multirow{2}{*}{Maintain}&Comp
&Relation&\multirow{2}{*}{No}&Flat&Static&\multirow{2}{*}{Yes}&\multirow{2}{*}{
$\iota~\delta~\chi$}&\multirow{2}{*}{
Active}&\multirow{2}{*}{Yes}&\multirow{2}{*}{Yes}&\multirow{2}{*}{S}&\multirow{2}{*}{Yes}&\multirow{2}{*}{No}&\multirow{2}{*}{---}&\multirow{2}{*}{No}&\multirow{2}{*}{No}\\
&&&&Run&Logic&&Limited&Dynamic&&&&&&&&&&&\\\hline

\multirow{2}{*}{\bf
\cite{Chen1}}&\multirow{2}{*}{H}&\multirow{2}{*}{Yes}&Check&Comp.
&O-O&Class
&\multirow{2}{*}{Limited}&\multirow{2}{*}{Static}&\multirow{2}{*}{Yes}&\multirow{2}{*}{
$\iota~\delta$}&\multirow{2}{*}{
Active}&\multirow{2}{*}{Yes}&\multirow{2}{*}{No}&\multirow{2}{*}{SS}&\multirow{2}{*}{Yes}&\multirow{2}{*}{No}&\multirow{2}{*}{1-6,9}&\multirow{2}{*}{No}&\multirow{2}{*}{Yes}\\
&&&Maintain&Run&&Att.&&&&&&&&&&&&&\\\hline

\multirow{2}{*}{\bf
\cite{Con}}&\multirow{2}{*}{N}&\multirow{2}{*}{Yes}&\multirow{2}{*}{Maintain}&\multirow{2}{*}{Run}
&\multirow{2}{*}{Logic}&\multirow{2}{*}{Yes}&Flat&\multirow{2}{*}{Static}&\multirow{2}{*}{Yes}&\multirow{2}{*}{
$\iota~\delta$}&\multirow{2}{*}{
Unfold.}&\multirow{2}{*}{Yes}&\multirow{2}{*}{No}&\multirow{2}{*}{S}&\multirow{2}{*}{Yes}&\multirow{2}{*}{No}&\multirow{2}{*}{1-6,9}&Not&\multirow{2}{*}{Yes}\\
&&&&&&&Limited&&&&&&&&&&&proved&\\\hline

\multirow{2}{*}{\bf
\cite{Lu1}}&\multirow{2}{*}{N}&\multirow{2}{*}{Yes}&\multirow{2}{*}{Maintain}&\multirow{2}{*}{Run}
&\multirow{2}{*}{Logic}&\multirow{2}{*}{Yes}&\multirow{2}{*}{Limited}&\multirow{2}{*}{Static}&\multirow{2}{*}{No}&\multirow{2}{*}{
$\iota~\delta$}&\multirow{2}{*}{
Active}&\multirow{2}{*}{Yes}&\multirow{2}{*}{No}&\multirow{2}{*}{SS}&\multirow{2}{*}{Yes}&\multirow{2}{*}{No}&\multirow{2}{*}{1-6,7}&\multirow{2}{*}{Yes}&Not\\
&&&&&&&&&&&&&&&&&&&proved\\\hline

\multirow{2}{*}{\bf
\cite{Ten}}&\multirow{2}{*}{N}&\multirow{2}{*}{Yes}&\multirow{2}{*}{Maintain}&Comp
&\multirow{2}{*}{Logic}&\multirow{2}{*}{Yes}&\multirow{2}{*}{Yes}&Static&\multirow{2}{*}{Yes}&\multirow{2}{*}{
$\iota~\delta$}&\multirow{2}{*}{
SLDNF}&\multirow{2}{*}{Yes}&\multirow{2}{*}{No}&\multirow{2}{*}{S}&\multirow{2}{*}{Yes}&\multirow{2}{*}{No}&\multirow{2}{*}{1-6,9}&\multirow{2}{*}{Yes}&\multirow{2}{*}{Yes}\\
&&&&Run&&&&Dynamic&&&&&&&&&&&\\\hline

\multirow{2}{*}{\bf
\cite{Sch}}&\multirow{2}{*}{S}&\multirow{2}{*}{Yes}&\multirow{2}{*}{Maintain}&\multirow{2}{*}{Comp}
&\multirow{2}{*}{Logic}&\multirow{2}{*}{No}&Flat&\multirow{2}{*}{Static}&\multirow{2}{*}{Yes}&\multirow{2}{*}{
$\iota~\delta$}&
predef&\multirow{2}{*}{---}&\multirow{2}{*}{Yes}&\multirow{2}{*}{G}&\multirow{2}{*}{No}&\multirow{2}{*}{Yes}&\multirow{2}{*}{---}&\multirow{2}{*}{No}&Not\\
&&&&&&&Limited&&&&Programs&&&&&&&&proved\\\hline

\end{tabular}
\end{turn}
\begin{turn}{180}
\text{\bf Tab. 1.\rm~Summary of view-update and integrity constraint
with our axiomatic method}
\end{turn}
\begin{turn}{180}\textbf{Appendix B}\end{turn}
\end{sidewaystable}

\begin{sidewaystable}
\centering
\begin{tabular}{|l|l|l|l|l|l|l|l|l|l|l|l|l|l|l|l|l|l|l|l|}\hline
\multirow{3}{*}{\bf Method}&\multicolumn{4}{c}{\bf Problem}&
\multicolumn{4}{|c|}{\bf Database
  schema}&\multicolumn{2}{c}{\bf Update req.}&\multicolumn{3}{|c|}{\bf Mechanism}&\multicolumn{3}{c}{\bf Update Change}&
\multicolumn{3}{|c|}{\bf Solutions}\\
\cline{2-20}

 &\multirow{2}{*}{Type}&View&IC&Run/&Def.&\multirow{2}{*}{View}&IC&Kind of
&\multirow{2}{*} {Mul.}&Update&Tech-&Base&User&\multirow{2}{*}{Type}&~Base~&View&\multirow{2}{*}{Axiom}&\multirow{2}{*}{Sound.}&\multirow{2}{*}{Complete.}\\
&&Update&Enforce.&Comp.&Lang.&&def.&IC&&Operat.&nique&Facts&Part.&&facts&def.&&&\\\hline

\multirow{2}{*}{\bf
\cite{Sta}}&\multirow{2}{*}{N}&\multirow{2}{*}{No}&\multirow{2}{*}{Maintain}&\multirow{2}{*}{Comp}
&\multirow{2}{*}{Logic}&\multirow{2}{*}{Yes}&\multirow{2}{*}{Limited}&\multirow{2}{*}{Static}&\multirow{2}{*}{Yes}&\multirow{2}{*}{
$\iota~\delta~\chi$}&
Predef&\multirow{2}{*}{Yes}&\multirow{2}{*}{No}&\multirow{2}{*}{S}&\multirow{2}{*}{Yes}&\multirow{2}{*}{No}&\multirow{2}{*}{1-6,7}&\multirow{2}{*}{Yes}&\multirow{2}{*}{No}\\
&&&&&&&&&&&Program&&&&&&&&\\\hline

\multirow{2}{*}{\bf
\cite{Arav2}}&\multirow{2}{*}{H}&\multirow{2}{*}{Yes}&\multirow{2}{*}{Check}&\multirow{2}{*}{Run}
&\multirow{2}{*}{Logic}&\multirow{2}{*}{Yes}&\multirow{2}{*}{Limited}&\multirow{2}{*}{Static}&\multirow{2}{*}{Yes}&\multirow{2}{*}{
$\iota~\delta$}&\multirow{2}{*}{
SLD}&\multirow{2}{*}{Yes}&\multirow{2}{*}{No}&\multirow{2}{*}{S}&\multirow{2}{*}{Yes}&\multirow{2}{*}{No}&\multirow{2}{*}{1-6,9}&\multirow{2}{*}{Yes}&\multirow{2}{*}{Yes}\\
&&&&&&&&&&&&&&&&&&&\\\hline

\multirow{2}{*}{\bf
\cite{Dec}}&\multirow{2}{*}{N}&\multirow{2}{*}{Yes}&\multirow{2}{*}{Maintain}&\multirow{2}{*}{Run}
&\multirow{2}{*}{Logic}&\multirow{2}{*}{Yes}&\multirow{2}{*}{Yes}&\multirow{2}{*}{Static}&\multirow{2}{*}{Yes}&\multirow{2}{*}{
$\iota~\delta$}&\multirow{2}{*}{
SLDNF}&\multirow{2}{*}{No}&\multirow{2}{*}{No}&\multirow{2}{*}{S}&\multirow{2}{*}{Yes}&\multirow{2}{*}{No}&\multirow{2}{*}{1-6,7}&\multirow{2}{*}{No}&Not\\
&&&&&&&&&&&&&&&&&&&Proved\\\hline

\multirow{2}{*}{\bf
\cite{Lobo}}&\multirow{2}{*}{N}&\multirow{2}{*}{Yes}&\multirow{2}{*}{Maintain}&\multirow{2}{*}{Run}
&\multirow{2}{*}{Logic}&\multirow{2}{*}{Yes}&Flat&\multirow{2}{*}{Static}&\multirow{2}{*}{Yes}&\multirow{2}{*}{
$\iota~\delta$}&\multirow{2}{*}{
Unfold}&\multirow{2}{*}{No}&\multirow{2}{*}{Yes}&\multirow{2}{*}{G}&\multirow{2}{*}{Yes}&\multirow{2}{*}{No}&\multirow{2}{*}{1-6,7}&Not&\multirow{2}{*}{No}\\
&&&&&&&Limited&&&&&&&&&&&proved&\\\hline

\multirow{2}{*}{\bf
\cite{Yang}}&\multirow{2}{*}{H}&\multirow{2}{*}{No}&\multirow{2}{*}{Maintain}&Comp.
&\multirow{2}{*}{Relation}&\multirow{2}{*}{Yes}&\multirow{2}{*}{Limited}&Static&\multirow{2}{*}{Yes}&\multirow{2}{*}{
$\iota~\delta~\chi$}&\multirow{2}{*}{
Unfold}&\multirow{2}{*}{Yes}&\multirow{2}{*}{No}&\multirow{2}{*}{S}&\multirow{2}{*}{Yes}&\multirow{2}{*}{No}&\multirow{2}{*}{1-6,7}&Not&Not\\
&&&&Run&&&&Dynamic&&&&&&&&&&proved&proved\\\hline

\multirow{2}{*}{\bf
\cite{Maa}}&\multirow{2}{*}{N}&\multirow{2}{*}{No}&Maintain&Comp
&\multirow{2}{*}{Logic}&\multirow{2}{*}{No}&Flat&Static&\multirow{2}{*}{Yes}&\multirow{2}{*}{
$\iota~\delta$}&\multirow{2}{*}{
Active}&\multirow{2}{*}{Yes}&\multirow{2}{*}{No}&\multirow{2}{*}{G}&\multirow{2}{*}{No}&\multirow{2}{*}{No}&\multirow{2}{*}{---}&\multirow{2}{*}{No}&\multirow{2}{*}{No}\\
&&&Restore&Run&&&Limited&Dynamic&&&&&&&&&&&\\\hline

\multirow{2}{*}{\bf
\cite{Sch1}}&\multirow{2}{*}{N}&\multirow{2}{*}{No}&\multirow{2}{*}{Maintain}&Comp
&\multirow{2}{*}{Relation}&\multirow{2}{*}{No}&Flat&\multirow{2}{*}{Static}&\multirow{2}{*}{Yes}&\multirow{2}{*}{
$\iota~\delta$}&\multirow{2}{*}{
Active}&\multirow{2}{*}{Yes}&\multirow{2}{*}{No}&\multirow{2}{*}{S}&\multirow{2}{*}{No}&\multirow{2}{*}{No}&\multirow{2}{*}{---}&\multirow{2}{*}{No}&\multirow{2}{*}{No}\\
&&&&Run&&&Limited&&&&&&&&&&&&\\\hline

\multirow{2}{*}{\bf
\cite{Lu}}&\multirow{2}{*}{N}&\multirow{2}{*}{Yes}&\multirow{2}{*}{Check}&\multirow{2}{*}{Run}
&\multirow{2}{*}{Logic}&\multirow{2}{*}{Yes}&\multirow{2}{*}{Limited}&\multirow{2}{*}{Static}&\multirow{2}{*}{Yes}&\multirow{2}{*}{
$\iota~\delta$}&\multirow{2}{*}{
SLD}&\multirow{2}{*}{Yes}&\multirow{2}{*}{No}&\multirow{2}{*}{S}&\multirow{2}{*}{Yes}&\multirow{2}{*}{No}&\multirow{2}{*}{1-6,9}&\multirow{2}{*}{Yes}&\multirow{2}{*}{Yes}\\
&&&&&&&&&&&&&&&&&&&\\\hline

\multirow{2}{*}{\bf
\cite{Agr}}&\multirow{2}{*}{O}&\multirow{2}{*}{No}&\multirow{2}{*}{Maintain}&\multirow{2}{*}{Run}
&\multirow{2}{*}{Logic}&\multirow{2}{*}{Yes}&\multirow{2}{*}{Limited}&\multirow{2}{*}{Static}&\multirow{2}{*}{Yes}&\multirow{2}{*}{
$\iota~\delta$}&\multirow{2}{*}{
---}&\multirow{2}{*}{Yes}&\multirow{2}{*}{No}&\multirow{2}{*}{S}&\multirow{2}{*}{Yes}&\multirow{2}{*}{No}&\multirow{2}{*}{---}&\multirow{2}{*}{No}&\multirow{2}{*}{No}\\
&&&&&&&&&&&&&&&&&&&\\\hline

\multirow{2}{*}{\bf
\cite{Sch2}}&\multirow{2}{*}{N}&\multirow{2}{*}{No}&\multirow{2}{*}{Maintain}&\multirow{2}{*}{Comp}
&\multirow{2}{*}{Relation}&\multirow{2}{*}{No}&\multirow{2}{*}{Limited}&\multirow{2}{*}{Static}&\multirow{2}{*}{Yes}&\multirow{2}{*}{
$\iota~\delta$}&Predef&\multirow{2}{*}{No}&\multirow{2}{*}{No}&\multirow{2}{*}{G}&\multirow{2}{*}{No}&\multirow{2}{*}{No}&\multirow{2}{*}{---}&\multirow{2}{*}{No}&\multirow{2}{*}{No}\\
&&&&&&&&&&&Program&&&&&&&&\\\hline

\multirow{2}{*}{\bf
\cite{Hal}}&\multirow{2}{*}{N}&\multirow{2}{*}{No}&\multirow{2}{*}{Maintain}&\multirow{2}{*}{Comp}
&\multirow{2}{*}{Logic}&\multirow{2}{*}{Yes}&\multirow{2}{*}{Limited}&\multirow{2}{*}{Static}&\multirow{2}{*}{Yes}&\multirow{2}{*}{
$\iota~\delta$}&\multirow{2}{*}{
---}&\multirow{2}{*}{No}&\multirow{2}{*}{No}&\multirow{2}{*}{S}&\multirow{2}{*}{No}&\multirow{2}{*}{No}&\multirow{2}{*}{---}&\multirow{2}{*}{No}&\multirow{2}{*}{No}\\
&&&&&&&&&&&&&&&&&&&\\\hline

\multirow{2}{*}{\bf
\cite{Chi}}&\multirow{2}{*}{N}&\multirow{2}{*}{No}&\multirow{2}{*}{Maintain}&Comp.
&\multirow{2}{*}{Relation}&\multirow{2}{*}{Yes}&\multirow{2}{*}{Limited}&Static&\multirow{2}{*}{Yes}&\multirow{2}{*}{
$\iota~\delta$}&\multirow{2}{*}{
---}&\multirow{2}{*}{Yes}&\multirow{2}{*}{No}&\multirow{2}{*}{S}&\multirow{2}{*}{Yes}&\multirow{2}{*}{No}&\multirow{2}{*}{---}&Not&Not\\
&&&&Run&&&&Dynamic&&&&&&&&&&proved&proved\\\hline

\multirow{2}{*}{\bf
\cite{Dom}}&\multirow{2}{*}{H}&\multirow{2}{*}{Yes}&\multirow{2}{*}{Check}&\multirow{2}{*}{Run}
&\multirow{2}{*}{Logic}&\multirow{2}{*}{Yes}&\multirow{2}{*}{Yes}&\multirow{2}{*}{Static}&\multirow{2}{*}{Yes}&\multirow{2}{*}{
$\iota~\delta$}&
Predef&\multirow{2}{*}{Yes}&\multirow{2}{*}{No}&\multirow{2}{*}{S}&\multirow{2}{*}{Yes}&\multirow{2}{*}{No}&\multirow{2}{*}{1-6,7}&\multirow{2}{*}{Yes}&Not\\
&&&&&&&&&&&Programs&&&&&&&&proved\\\hline

\multirow{2}{*}{\bf
\cite{Heg1}}&\multirow{2}{*}{O}&\multirow{2}{*}{Yes}&\multirow{2}{*}{Check}&\multirow{2}{*}{Run}
&\multirow{2}{*}{Relation}&\multirow{2}{*}{Yes}&\multirow{2}{*}{Limited}&\multirow{2}{*}{Static}&\multirow{2}{*}{No}&\multirow{2}{*}{
$\iota~\delta$}&\multirow{2}{*}{
Unfold}&\multirow{2}{*}{Yes}&\multirow{2}{*}{No}&\multirow{2}{*}{S}&\multirow{2}{*}{Yes}&\multirow{2}{*}{No}&\multirow{2}{*}{1-6,9}&\multirow{2}{*}{Yes}&\multirow{2}{*}{Yes}\\
&&&&&&&&&&&&&&&&&&&\\\hline

\multirow{2}{*}{\bf
\cite{Baue}}&\multirow{2}{*}{O}&\multirow{2}{*}{Yes}&\multirow{2}{*}{Check}&\multirow{2}{*}{Run}
&\multirow{2}{*}{Relation}&\multirow{2}{*}{Yes}&\multirow{2}{*}{Limited}&\multirow{2}{*}{Static}&\multirow{2}{*}{No}&\multirow{2}{*}{
$\iota~\delta$}&\multirow{2}{*}{
Unfold}&\multirow{2}{*}{Yes}&\multirow{2}{*}{No}&\multirow{2}{*}{S}&\multirow{2}{*}{Yes}&\multirow{2}{*}{No}&\multirow{2}{*}{1-6,9}&\multirow{2}{*}{Yes}&\multirow{2}{*}{Yes}\\
&&&&&&&&&&&&&&&&&&&\\\hline

\multirow{2}{*}{\bf
\cite{Far}}&\multirow{2}{*}{N}&\multirow{2}{*}{Yes}&\multirow{2}{*}{Check}&\multirow{2}{*}{Run}
&\multirow{2}{*}{Logic}&\multirow{2}{*}{Yes}&\multirow{2}{*}{Yes}&\multirow{2}{*}{Static}&\multirow{2}{*}{Yes}&\multirow{2}{*}{
$\iota~\delta$}&\multirow{2}{*}{
SLDNF}&\multirow{2}{*}{Yes}&\multirow{2}{*}{No}&\multirow{2}{*}{S}&\multirow{2}{*}{Yes}&\multirow{2}{*}{No}&\multirow{2}{*}{1-6,9}&\multirow{2}{*}{Yes}&\multirow{2}{*}{Yes}\\
&&&&&&&&&&&&&&&&&&&\\\hline

\multirow{2}{*}{\bf
\cite{Saha}}&\multirow{2}{*}{N}&\multirow{2}{*}{No}&\multirow{2}{*}{Maintain}&\multirow{2}{*}{Run}
&\multirow{2}{*}{Logic}&\multirow{2}{*}{Yes}&\multirow{2}{*}{Limited}&\multirow{2}{*}{Static}&\multirow{2}{*}{Yes}&\multirow{2}{*}{
$\iota~\delta$}&\multirow{2}{*}{
Predef}&\multirow{2}{*}{Yes}&\multirow{2}{*}{No}&\multirow{2}{*}{S}&\multirow{2}{*}{Yes}&\multirow{2}{*}{No}&\multirow{2}{*}{1-6,7}&\multirow{2}{*}{Yes}&Not\\
&&&&&&&&&&&Programs&&&&&&&&proved\\\hline

\multirow{2}{*}{\bf
\cite{Hor}}&\multirow{2}{*}{O}&\multirow{2}{*}{No}&\multirow{2}{*}{Maintain}&\multirow{2}{*}{Comp}
&\multirow{2}{*}{Relation}&\multirow{2}{*}{Yes}&\multirow{2}{*}{Limited}&\multirow{2}{*}{Static}&\multirow{2}{*}{Yes}&\multirow{2}{*}{
$\iota~\delta$}&\multirow{2}{*}{
---}&\multirow{2}{*}{Yes}&\multirow{2}{*}{No}&\multirow{2}{*}{S}&\multirow{2}{*}{Yes}&\multirow{2}{*}{No}&\multirow{2}{*}{---}&\multirow{2}{*}{No}&\multirow{2}{*}{No}\\
&&&&&&&&&&&&&&&&&&&\\\hline

\end{tabular}
\end{sidewaystable}

\begin{sidewaystable}
\centering
\begin{turn}{180}
\begin{tabular}{|l|l|l|l|l|l|l|l|l|l|l|l|l|l|l|l|l|l|l|l|}\hline
\multirow{3}{*}{\bf Method}&\multicolumn{4}{c}{\bf Problem}&
\multicolumn{4}{|c|}{\bf Database
  schema}&\multicolumn{2}{c}{\bf Update req.}&\multicolumn{3}{|c|}{\bf Mechanism}&\multicolumn{3}{c}{\bf Update Change}&
\multicolumn{3}{|c|}{\bf Solutions}\\
\cline{2-20}

 &\multirow{2}{*}{Type}&View&IC&Run/&Def.&\multirow{2}{*}{View}&IC&Kind of
&\multirow{2}{*} {Mul.}&Update&Tech-&Base&User&\multirow{2}{*}{Type}&~Base~&View&\multirow{2}{*}{Axiom}&\multirow{2}{*}{Sound.}&\multirow{2}{*}{Complete.}\\
&&Update&Enforce.&Comp.&Lang.&&def.&IC&&Operat.&nique&Facts&Part.&&facts&def.&&&\\\hline

\multirow{2}{*}{\bf
\cite{Sak}}&\multirow{2}{*}{N}&\multirow{2}{*}{Yes}&\multirow{2}{*}{Check}&\multirow{2}{*}{Run}
&\multirow{2}{*}{Logic}&\multirow{2}{*}{No}&\multirow{2}{*}{Limited}&\multirow{2}{*}{Static}&\multirow{2}{*}{Yes}&\multirow{2}{*}{
$\iota~\delta$}&\multirow{2}{*}{
SLDNF}&\multirow{2}{*}{Yes}&\multirow{2}{*}{No}&\multirow{2}{*}{S}&\multirow{2}{*}{Yes}&\multirow{2}{*}{No}&\multirow{2}{*}{1-6,9}&\multirow{2}{*}{Yes}&\multirow{2}{*}{Yes}\\
&&&&&&&&&&&&&&&&&&&\\\hline

\multirow{2}{*}{\bf
\cite{Far1}}&\multirow{2}{*}{N}&\multirow{2}{*}{Yes}&\multirow{2}{*}{Check}&\multirow{2}{*}{Run}
&\multirow{2}{*}{Logic}&\multirow{2}{*}{Yes}&\multirow{2}{*}{Yes}&\multirow{2}{*}{Static}&\multirow{2}{*}{No}&\multirow{2}{*}{
$\iota~\delta$}&\multirow{2}{*}{
SLDNF}&\multirow{2}{*}{No}&\multirow{2}{*}{No}&\multirow{2}{*}{S}&\multirow{2}{*}{Yes}&\multirow{2}{*}{No}&\multirow{2}{*}{1-6,9}&Not&\multirow{2}{*}{No}\\
&&&&&&&&&&&&&&&&&&&proved\\\hline

\multirow{2}{*}{\bf
\cite{Mar}}&\multirow{2}{*}{N}&\multirow{2}{*}{Yes}&Check&\multirow{2}{*}{Run}
&\multirow{2}{*}{Logic}&\multirow{2}{*}{Yes}&\multirow{2}{*}{Yes}&\multirow{2}{*}{Static}&\multirow{2}{*}{No}&\multirow{2}{*}{
$\iota~\delta~\chi$}&\multirow{2}{*}{
SLD}&\multirow{2}{*}{Yes}&\multirow{2}{*}{No}&\multirow{2}{*}{S}&\multirow{2}{*}{Yes}&\multirow{2}{*}{No}&\multirow{2}{*}{---}&\multirow{2}{*}{No}&\multirow{2}{*}{No}\\
&&&Maintain&&&&&&&&&&&&&&&&\\\hline

\multirow{2}{*}{\bf
\cite{Bra}}&\multirow{2}{*}{N}&\multirow{2}{*}{Yes}&Check&Run
&\multirow{2}{*}{Logic}&\multirow{2}{*}{Yes}&\multirow{2}{*}{Yes}&\multirow{2}{*}{Static}&\multirow{2}{*}{No}&\multirow{2}{*}{
$\iota~\delta~\chi$}&\multirow{2}{*}{
SLD}&\multirow{2}{*}{Yes}&\multirow{2}{*}{No}&\multirow{2}{*}{SS}&\multirow{2}{*}{Yes}&\multirow{2}{*}{No}&\multirow{2}{*}{1-6,7}&\multirow{2}{*}{Yes}&\multirow{2}{*}{Not}\\
&&&Maintain&Comp&&&&&&&&&&&&&&&proved\\\hline

\multirow{2}{*}{\bf
\cite{Chris2}}&\multirow{2}{*}{N}&\multirow{2}{*}{Yes}&Check&\multirow{2}{*}{Run}
&\multirow{2}{*}{Logic}&\multirow{2}{*}{Yes}&\multirow{2}{*}{Yes}&Static&\multirow{2}{*}{Yes}&\multirow{2}{*}{
$\iota~\delta~\chi$}&\multirow{2}{*}{
Predef}&\multirow{2}{*}{Yes}&\multirow{2}{*}{No}&\multirow{2}{*}{S}&\multirow{2}{*}{Yes}&\multirow{2}{*}{No}&\multirow{2}{*}{1-6,9}&\multirow{2}{*}{Yes}&\multirow{2}{*}{Yes}\\
&&&Maintain&&&&&Dynamic&&&Program&&&&&&&&\\\hline

\multirow{2}{*}{\bf
\cite{Car}}&\multirow{2}{*}{N}&\multirow{2}{*}{Yes}&\multirow{2}{*}{Check}&\multirow{2}{*}{Comp}
&\multirow{2}{*}{Logic}&\multirow{2}{*}{Yes}&\multirow{2}{*}{Yes}&\multirow{2}{*}{Dynamic}&\multirow{2}{*}{Yes}&\multirow{2}{*}{
$\iota~\delta$}&\multirow{2}{*}{
Predef}&\multirow{2}{*}{Yes}&\multirow{2}{*}{No}&\multirow{2}{*}{S}&\multirow{2}{*}{Yes}&\multirow{2}{*}{No}&\multirow{2}{*}{---}&Not&\multirow{2}{*}{No}\\
&&&&&&&&&&&Programs&&&&&&&Proved&proved\\\hline

\multirow{2}{*}{\bf
\cite{Chris1}}&\multirow{2}{*}{N}&\multirow{2}{*}{Yes}&Check&\multirow{2}{*}{Run}
&\multirow{2}{*}{Logic}&\multirow{2}{*}{Yes}&\multirow{2}{*}{Yes}&Static&\multirow{2}{*}{Yes}&\multirow{2}{*}{
$\iota~\delta~\chi$}&\multirow{2}{*}{
Predef}&\multirow{2}{*}{Yes}&\multirow{2}{*}{No}&\multirow{2}{*}{S}&\multirow{2}{*}{Yes}&\multirow{2}{*}{No}&\multirow{2}{*}{1-6,9}&\multirow{2}{*}{Yes}&\multirow{2}{*}{Yes}\\
&&&Maintain&&&&&Dynamic&&&Program&&&&&&&&\\\hline

\multirow{2}{*}{\bf
\cite{Coe}}&\multirow{2}{*}{N}&\multirow{2}{*}{No}&\multirow{2}{*}{Maintain}&\multirow{2}{*}{Comp}
&\multirow{2}{*}{Logic}&\multirow{2}{*}{Yes}&\multirow{2}{*}{No}&\multirow{2}{*}{---}&\multirow{2}{*}{Yes}&\multirow{2}{*}{
$\iota~\delta$}&\multirow{2}{*}{
---}&\multirow{2}{*}{Yes}&\multirow{2}{*}{No}&\multirow{2}{*}{S}&\multirow{2}{*}{Yes}&\multirow{2}{*}{No}&\multirow{2}{*}{---}&\multirow{2}{*}{No}&\multirow{2}{*}{No}\\
&&&&&&&&&&&&&&&&&&&\\\hline

\multirow{2}{*}{\bf
\cite{Zho}}&\multirow{2}{*}{N}&\multirow{2}{*}{No}&\multirow{2}{*}{Maintain}&\multirow{2}{*}{Run}
&\multirow{2}{*}{Relation}&\multirow{2}{*}{Yes}&\multirow{2}{*}{No}&\multirow{2}{*}{---}&\multirow{2}{*}{Yes}&\multirow{2}{*}{
$\iota~\delta~\chi$}&\multirow{2}{*}{
Unfold}&\multirow{2}{*}{Yes}&\multirow{2}{*}{No}&\multirow{2}{*}{SS}&\multirow{2}{*}{No}&\multirow{2}{*}{No}&\multirow{2}{*}{---}&Not&Not\\
&&&&&&&&&&&&&&&&&&proved&proved\\\hline

\multirow{2}{*}{\bf
\cite{Heg}}&\multirow{2}{*}{O}&\multirow{2}{*}{No}&\multirow{2}{*}{Maintain}&Comp.
&\multirow{2}{*}{Logic}&\multirow{2}{*}{Yes}&\multirow{2}{*}{No}&\multirow{2}{*}{---}&\multirow{2}{*}{Yes}&\multirow{2}{*}{
$\iota~\delta$}&\multirow{2}{*}{
----}&\multirow{2}{*}{Yes}&\multirow{2}{*}{No}&\multirow{2}{*}{G}&\multirow{2}{*}{No}&\multirow{2}{*}{No}&\multirow{2}{*}{---}&\multirow{2}{*}{Yes}&Not\\
&&&&Run&&&&&&&&&&&&&&&proved\\\hline

\multirow{2}{*}{\bf
\cite{Beh}}&\multirow{2}{*}{S}&\multirow{2}{*}{Yes}&\multirow{2}{*}{Check}&\multirow{2}{*}{Run}
&\multirow{2}{*}{Logic}&\multirow{2}{*}{Yes}&Flat&\multirow{2}{*}{Static}&\multirow{2}{*}{Yes}&\multirow{2}{*}{
$\iota~\delta$}&\multirow{2}{*}{
SLDNF}&\multirow{2}{*}{Yes}&\multirow{2}{*}{No}&\multirow{2}{*}{S}&\multirow{2}{*}{Yes}&\multirow{2}{*}{No}&\multirow{2}{*}{1-6,9}&\multirow{2}{*}{Yes}&Not\\
&&&&&&&Limited&&&&&&&&&&&&proved\\\hline

\multirow{2}{*}{\bf
\cite{Dom1}}&\multirow{2}{*}{N}&\multirow{2}{*}{Yes}&\multirow{2}{*}{Check}&\multirow{2}{*}{Run}
&\multirow{2}{*}{Logic}&\multirow{2}{*}{Yes}&\multirow{2}{*}{Yes}&\multirow{2}{*}{Static}&\multirow{2}{*}{Yes}&\multirow{2}{*}{
$\iota~\delta$}&\multirow{2}{*}{
---}&\multirow{2}{*}{Yes}&\multirow{2}{*}{No}&\multirow{2}{*}{S}&\multirow{2}{*}{Yes}&\multirow{2}{*}{No}&\multirow{2}{*}{---}&\multirow{2}{*}{No}&\multirow{2}{*}{No}\\
&&&&&&&&&&&&&&&&&&&\\\hline

\multirow{2}{*}{\bf
\cite{Cha}}&\multirow{2}{*}{O}&\multirow{2}{*}{No}&\multirow{2}{*}{Maintain}&\multirow{2}{*}{Run}
&\multirow{2}{*}{Relation}&\multirow{2}{*}{Yes}&\multirow{2}{*}{No}&\multirow{2}{*}{Static}&\multirow{2}{*}{Yes}&\multirow{2}{*}{
$\iota~\delta$}&\multirow{2}{*}{
SLD}&\multirow{2}{*}{Yes}&\multirow{2}{*}{Yes}&\multirow{2}{*}{G}&\multirow{2}{*}{No}&\multirow{2}{*}{No}&\multirow{2}{*}{---}&Not&Not\\
&&&&&&&&&&&&&&&&&&proved&proved\\\hline

\multirow{2}{*}{\bf
\cite{Bell}}&\multirow{2}{*}{O}&\multirow{2}{*}{No}&\multirow{2}{*}{Maintain}&\multirow{2}{*}{Comp}
&\multirow{2}{*}{Relation}&\multirow{2}{*}{Yes}&\multirow{2}{*}{No}&\multirow{2}{*}{Static}&\multirow{2}{*}{Yes}&\multirow{2}{*}{
$\iota~\delta~\chi$}&\multirow{2}{*}{
---}&\multirow{2}{*}{Yes}&\multirow{2}{*}{No}&\multirow{2}{*}{SS}&\multirow{2}{*}{Yes}&\multirow{2}{*}{No}&\multirow{2}{*}{---}&\multirow{2}{*}{No}&\multirow{2}{*}{No}\\
&&&&&&&&&&&&&&&&&&&\\\hline

\multirow{2}{*}{\bf
\cite{Alex}}&\multirow{2}{*}{O}&\multirow{2}{*}{No}&\multirow{2}{*}{Maintain}&Comp.
&\multirow{2}{*}{Relation}&\multirow{2}{*}{No}&\multirow{2}{*}{Limited}&Static&\multirow{2}{*}{Yes}&\multirow{2}{*}{
$\iota~\delta$}&\multirow{2}{*}{
---}&\multirow{2}{*}{Yes}&\multirow{2}{*}{No}&\multirow{2}{*}{G}&\multirow{2}{*}{Yes}&\multirow{2}{*}{No}&\multirow{2}{*}{---}&\multirow{2}{*}{No}&\multirow{2}{*}{No}\\
&&&&Run&&&&Dynamic&&&&&&&&&&&\\\hline

\multirow{2}{*}{\bf
\cite{Mam1}}&\multirow{2}{*}{N}&\multirow{2}{*}{No}&\multirow{2}{*}{Maintain}&\multirow{2}{*}{Comp}
&\multirow{2}{*}{Relation}&\multirow{2}{*}{No}&\multirow{2}{*}{Yes}&\multirow{2}{*}{Static}&\multirow{2}{*}{Yes}&\multirow{2}{*}{
$\iota~\delta~\chi$}&\multirow{2}{*}{
Unfold}&\multirow{2}{*}{No}&\multirow{2}{*}{Yes}&\multirow{2}{*}{SS}&\multirow{2}{*}{No}&\multirow{2}{*}{No}&\multirow{2}{*}{---}&\multirow{2}{*}{No}&\multirow{2}{*}{No}\\
&&&&&&&&&&&&&&&&&&&\\\hline

\multirow{2}{*}{\bf
\cite{Sal}}&\multirow{2}{*}{N}&\multirow{2}{*}{No}&\multirow{2}{*}{Check}&\multirow{2}{*}{Comp}
&\multirow{2}{*}{Logic}&\multirow{2}{*}{No}&\multirow{2}{*}{Yes}&\multirow{2}{*}{Static}&\multirow{2}{*}{Yes}&\multirow{2}{*}{
$\iota~\delta$}&\multirow{2}{*}{
Active}&\multirow{2}{*}{Yes}&\multirow{2}{*}{No}&\multirow{2}{*}{G}&\multirow{2}{*}{Yes}&\multirow{2}{*}{No}&\multirow{2}{*}{---}&\multirow{2}{*}{No}&\multirow{2}{*}{No}\\
&&&&&&&&&&&&&&&&&&&\\\hline

\multirow{2}{*}{\bf
\cite{Del2}}&\multirow{2}{*}{N}&\multirow{2}{*}{Yes}&\multirow{2}{*}{Check}&\multirow{2}{*}{Run}
&\multirow{2}{*}{Logic}&\multirow{2}{*}{Yes}&\multirow{2}{*}{Yes}&\multirow{2}{*}{Static}&\multirow{2}{*}{Yes}&\multirow{2}{*}{
$\iota~\delta$}&\multirow{2}{*}{
SLD}&\multirow{2}{*}{Yes}&\multirow{2}{*}{No}&\multirow{2}{*}{S}&\multirow{2}{*}{Yes}&\multirow{2}{*}{No}&\multirow{2}{*}{1-6,9}&
\multirow{2}{*}{Yes}&\multirow{2}{*}{Yes}\\
&&&&&&&&&&&&&&&&&&&\\\hline

\end{tabular}
\end{turn}
\end{sidewaystable}

\section*{Acknowledgement}

The author acknowledges the support of RWTH Aachen, where he is
visiting scholar with an Erasmus Mundus External Cooperation Window
India4EU by the European Commission when the paper was written. I
would like to thanks Chandrabose Aravindan and Gerhard Lakemeyer
both my Indian and Germany PhD supervisor, give encourage to write
the paper.



\end{document}